\documentclass[lettersize,journal]{IEEEtran}
\usepackage{subcaption}
\usepackage{cite}
\usepackage{amsthm, amsmath,amsfonts}
\usepackage{algorithmic}
\usepackage{setspace}
\usepackage{algorithm}
\usepackage{multirow}
\usepackage{amssymb}
\usepackage{pifont}
\usepackage{booktabs}
\usepackage{enumerate}
\usepackage{array}
\usepackage{textcomp}
\usepackage{stfloats}
\usepackage{url}
\usepackage{verbatim}
\usepackage{graphicx}
\usepackage{xcolor}
\usepackage{cite}
\begin{document}

\theoremstyle{definition}
\newtheorem{definition}{Definition}

\title{Metapath-based Hyperbolic Contrastive Learning for Heterogeneous Graph Embedding}

\author{Jongmin Park, Seunghoon Han, Won-Yong Shin,~\IEEEmembership{Senior Member,~IEEE}, Sungsu Lim*,~\IEEEmembership{Member,~IEEE}
        % <-this % stops a space
\thanks{*Corresponding author.}}

\newcommand{\xmark}{{\color{red}\ding{55}}}
\newcommand{\cmark}{\color{green}\ding{51}}
\newcommand{\algname}{MHCL}
\newcommand{\Poincare}{Poincar\'e }
\newcommand{\Mobius}{M\"obius }
\newcommand{\jong}[1]{\textcolor{blue}{#1}}
\newcommand{\red}[1]{\textcolor{red}{#1}}

% The paper headers
%\markboth{Journal of \LaTeX\ Class Files,~Vol.~14, No.~8, August~2021}%
%{Shell \MakeLowercase{\textit{et al.}}: A Sample Article Using IEEEtran.cls for IEEE Journals}

% Remember, if you use this you must call \IEEEpubidadjcol in the second
% column for its text to clear the IEEEpubid mark.

\maketitle

\begin{abstract}
%Heterogeneous graphs, consisting of multiple types of nodes and links, are pervasive in real-world scenarios. Accordingly, the demand for effective methods to learn semantic information and complex structures within such graphs is increasing steadily. 
In heterogeneous graphs, a metapath can be defined as a sequence of node or link types, allowing the learning of both semantic information and structural properties. From a structural perspective, various hierarchical or power-law structures, each corresponding to a specific metapath, can be observed in real-world heterogeneous graphs.
Recent studies in heterogeneous graph embedding use hyperbolic space to capture such complex structures. The hyperbolic space, characterized by a constant negative curvature and exponentially expanding space, aligns well with the structural properties of heterogeneous graphs. However, although heterogeneous graphs inherently possess diverse power-law structures, most hyperbolic heterogeneous graph embedding models rely on a single hyperbolic space. This approach may fail to effectively capture the diverse power-law structures within heterogeneous graphs. 

To address this limitation, we propose a \textit{\underline{M}etapath-based \underline{H}yperbolic \underline{C}ontrastive \underline{L}earning framework} (\algname{}), which uses multiple hyperbolic spaces to capture diverse complex structures within heterogeneous graphs. Specifically, by learning each hyperbolic space to describe the distribution of complex structures corresponding to each metapath, it is possible to capture semantic information effectively. Since metapath embeddings represent distinct semantic information, preserving their discriminability is important when aggregating them to obtain node representations. Therefore, we use a contrastive learning approach to optimize \algname{} and improve the discriminability of metapath embeddings. In particular, our contrastive learning method minimizes the distance between embeddings of the same metapath and maximizes the distance between those of different metapaths in hyperbolic space, thereby improving the separability of metapath embeddings with distinct semantic information. We conduct comprehensive experiments to evaluate the effectiveness of \algname{}. The experimental results demonstrate that \algname{} outperforms state-of-the-art baselines in various graph machine learning tasks, effectively capturing the complex structures of heterogeneous graphs.
\end{abstract}


\begin{figure}[t!]
\centering
    \includegraphics[width=\columnwidth]{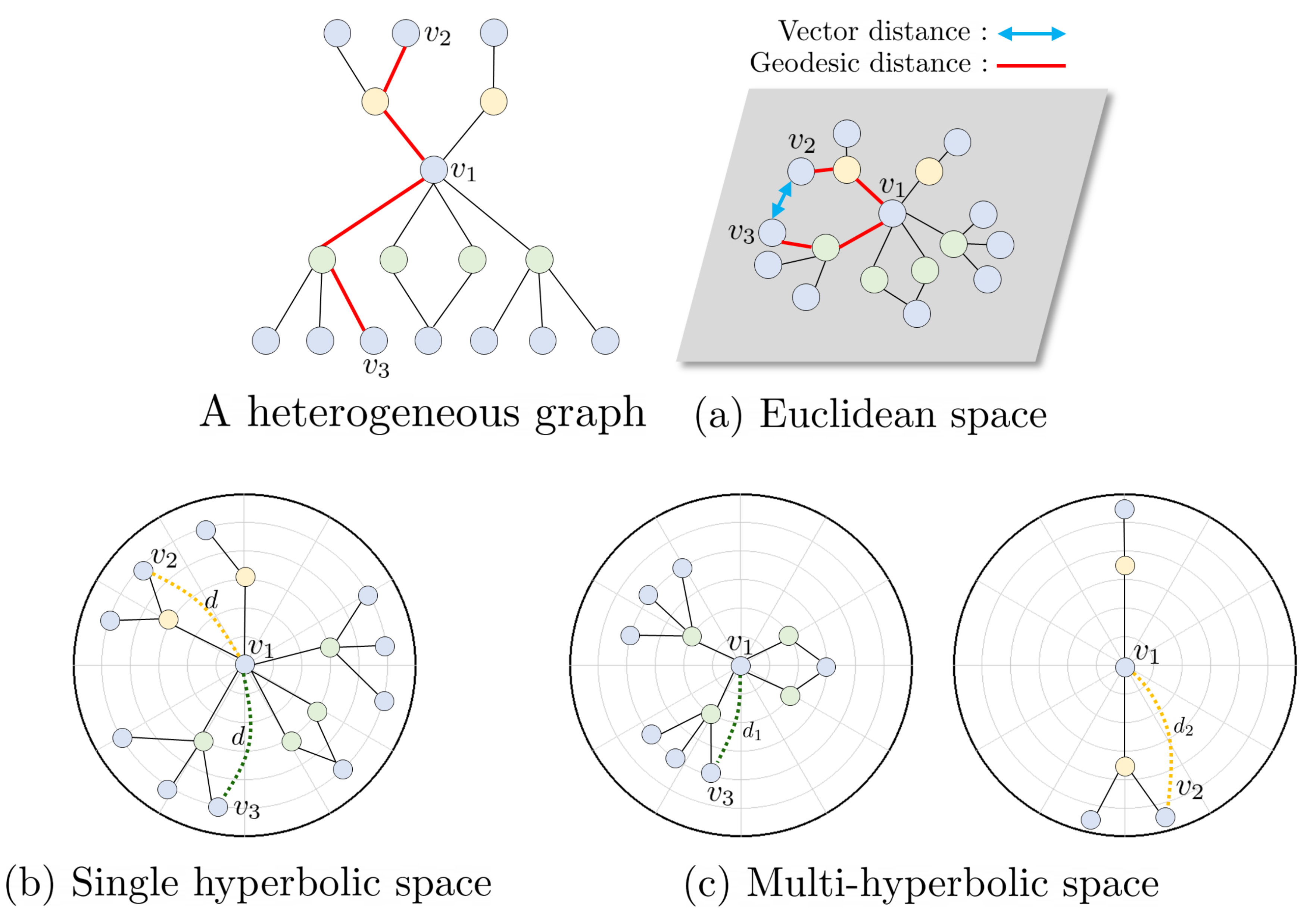}
    \caption{Examples of heterogeneous graph representations in various embedding spaces.}
    \label{figure:toy_example}
\end{figure}

\section{Introduction}
\IEEEPARstart{H}{eterogeneous graphs}, which consist of multiple types of nodes and links, have emerged powerful representations for modeling complex real-world scenarios, including social networks, bibliographic databases, and molecular structures. In heterogeneous graphs, we can capture both semantic and structural information from metapaths~\cite{sun2011pathsim}, which are defined as the ordered sequence of node and link types.

Accordingly, recent studies have focused on effectively learning semantic and structural information guided by metapaths to enhance the heterogeneous graph representation learning. HAN~\cite{wang2019heterogeneous} proposed heterogeneous Graph Neural Networks (GNNs) to aggregate information from metapath-based neighbors. GTN~\cite{yun2019graph} learns soft selection by utilizing graph transformer layers to learn metapaths. GraphMSE~\cite{li2021graphmse} proposed semantic feature space alignment, which aligns the metapath instance vector representations to facilitate automatic metapfath selection. HGT~\cite{hu2020heterogeneous} proposed node and link type-dependent attention mechanisms to capture the various types of relations within heterogeneous graphs. Simple-HGN~\cite{lv2021we} extended graph attention mechanisms by calculating attention scores through incorporating link-type information. 

Although existing methods have demonstrated significant effectiveness in heterogeneous graph representation learning, they may struggle to accurately capture complex structures such as hierarchical or power-law structures, due to their reliance on the Euclidean space for heterogeneous graph embedding. In heterogeneous graphs, such complex structures can be observed, and representing them in the Euclidean space can lead to distortions and limitations~\cite{pei2020curvature}. For example, as illustrated in Figure~\ref{figure:toy_example}(a), in the case of a hierarchical graph, distortion can occur when two nodes ($v_2$ and $v_3$) that are far apart of geodesic distance are represented in close proximity within the Euclidean space.

Some recent heterogeneous graph embedding models have addressed this challenge by using hyperbolic spaces as the embedding space. Unlike the Euclidean space, the hyperbolic space has a constant negative curvature and grows exponentially. Several studies~\cite{nickel2017poincare, balazevic2019multi, pan2021hyperbolic, peng2021hyperbolic, yang2022hyperbolic} argue that these inherent properties of the hyperbolic space provide a natural solution for effectively representing complex structures. HHNE~\cite{wang2019hyperbolic} proposed a shallow heterogeneous graph embedding model that utilizes a metapath-based random walk strategy to sample metapath instances and embeds them in the hyperbolic space. SHNE~\cite{li2023multi} proposed hyperbolic heterogeneous GNNs that sample simplicial complexes from heterogeneous graphs and use graph attention mechanisms in the hyperbolic space to learn multi-order relations within such graphs.
HHGAT~\cite{park2024hyperbolic} proposed hyperbolic graph attention networks to learn semantic information and complex structures by leveraging metapath instances within heterogeneous graphs. 

These studies~\cite{wang2019hyperbolic, li2023multi, park2024hyperbolic} demonstrate the effectiveness of the hyperbolic space in representing complex structures, such as hierarchical or power-law structures. However, despite their significant achievements, a critical question remains: `` Is it truly effective to represent heterogeneous graphs, which may include various complex structures based on multiple metapaths, within a single hyperbolic space?" In the hyperbolic space, the extent to which hyperbolic space grows exponentially is determined by the negative curvature. Conversely, we can interpret negative curvature as indicating the degree of power-law distribution in the hyperbolic space. Therefore, utilizing multiple hyperbolic spaces with distinct negative curvatures that represent each power-law structure within a heterogeneous graph would be more effective for heterogeneous graph representation learning.

\begin{figure*}[t!]
\captionsetup[subfigure]{justification=centering}
\centering
        \begin{minipage}[b]{0.33\textwidth}
                \centering
                \includegraphics[width=\linewidth]{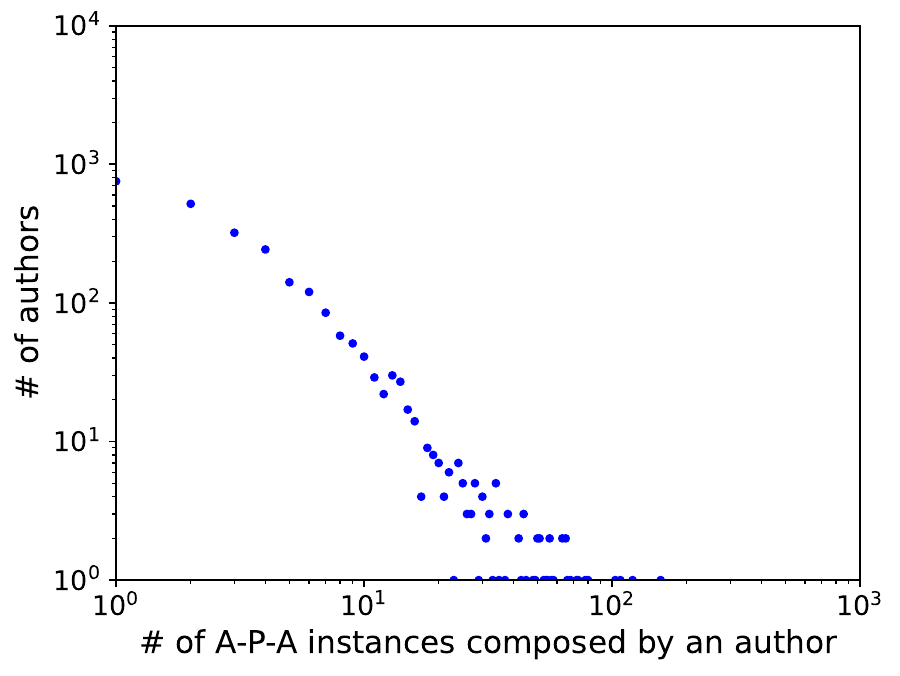}
                \subcaption{A-P-A in DBLP \\($\delta_{avg}$ : 0.7892)}
        \end{minipage}%
        \begin{minipage}[b]{0.33\textwidth}
                \centering
                \includegraphics[width=\linewidth]{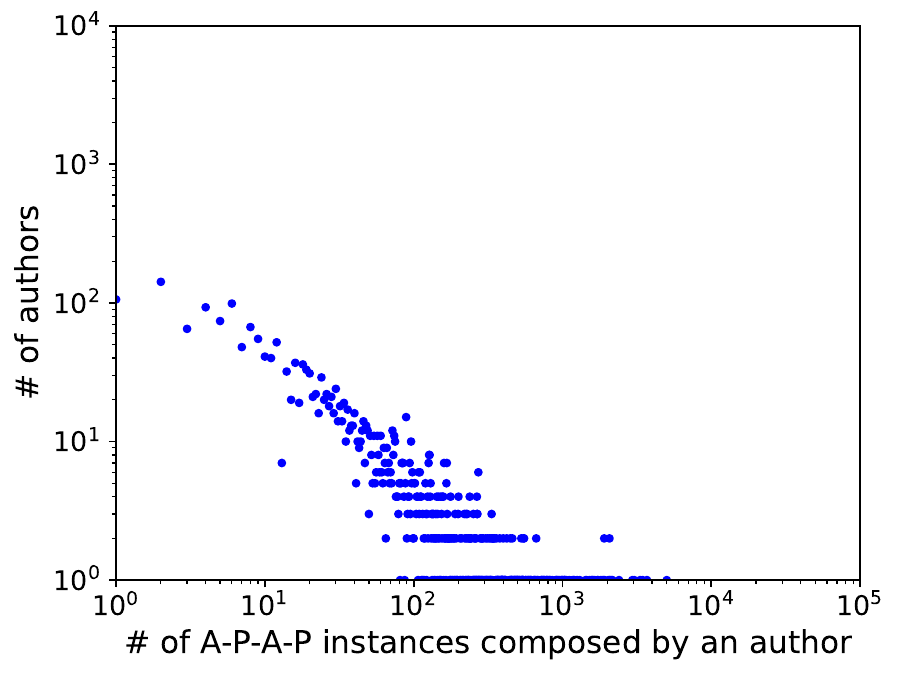}
                \subcaption{A-P-A-P in DBLP \\($\delta_{avg}$ : 0.7865)}
        \end{minipage}
        \begin{minipage}[b]{0.33\textwidth}
                \centering
                \includegraphics[width=\linewidth]{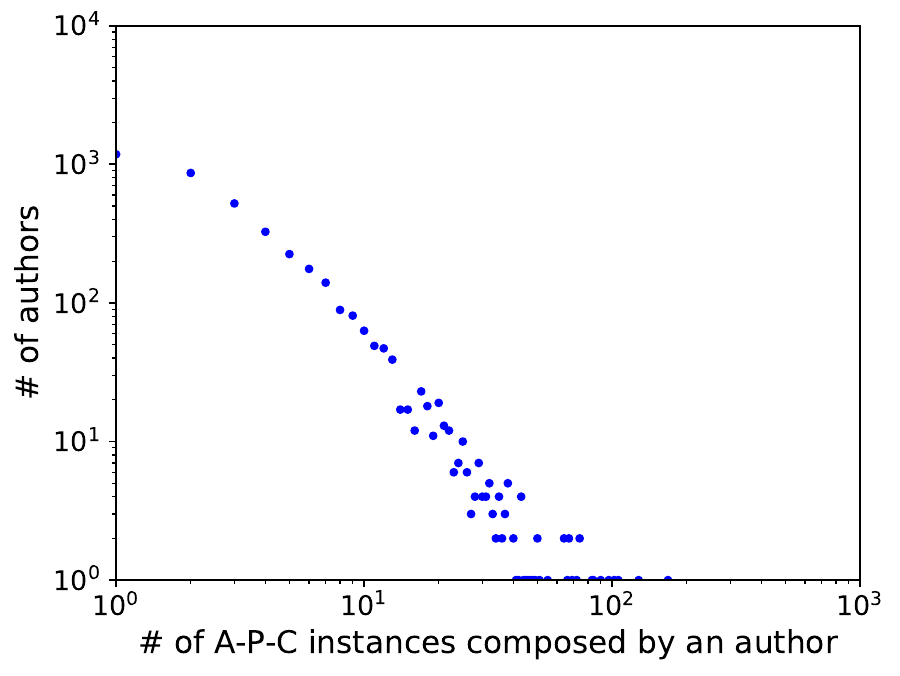}
                \subcaption{A-P-C in DBLP\\($\delta_{avg}$ : 0.3124)}
        \end{minipage}%
        \caption{Metapath instance distributions of some metapaths on the DBLP dataset. Here, $\delta$ represents Gromov's $\delta$-hyperbolicity of each metapath-based subgraph.}
        \label{fig:distributions}
        
\end{figure*}

For examples, as illustrated in Figure~\ref{figure:toy_example}(b), learning power-law structures with different degree distributions in a single hyperbolic space can lead to ineffective modeling of their distinct structural properties. Specifically, when two different metapaths, each corresponding to different degree distributions, are represented, their metapath-based neighbors are positioned at the same hyperbolic distance $d$ from the central target node $v_1$, failing to capture their structural differences. In contrast, as shown in Figure~\ref{figure:toy_example}(c), if structures with stronger power-law structures are represented in a hyperbolic space with steeper curvature, and those with weaker power-law structures are represented in a hyperbolic space with softer curvature, the structural properties of each metapath can be more effectively preserved. In this case, neighbors derived from different metapaths are positioned at distinct hyperbolic space $d_1$ and $d_2$ from the central node $v_1$, thereby capturing their structural differences.

Moreover, as illustrated in Figure~\ref{fig:distributions}, in real-world heterogeneous graphs, we can observe multiple power-law structures corresponding to specific metapaths that are similar but distinct. %\jong{As shown in Figure~\ref{fig:distributions}(a) and (c), the metapath A–P–A (Author–Paper–Author) corresponds to a dense co-authorship structure with a low branching factor, while A–P–C (Author–Paper–Conference) corresponds to a sparse, many-to-one structure with a high branching factor.} 
To quantify these differences, we calculate the average Gromov $\delta$-hyperbolicity~\cite{adcock2013tree, jonckheere2008scaled, narayan2011large} for each metapath-based subgraph, where lower $\delta$ values indicate that the structure of the subgraph tends to exhibit more hierarchical structures. %\jong{These structural differences have a direct impact on the curvature requirements of the hyperbolic space.}

%The average Gromov $\delta$-hyperbolicity allows for distinguishing distinct hierarchical structures in power-law structures with similar distributions, despite their apparent similarity.

\begin{figure*}[t!]
\centering
    \includegraphics[width=\textwidth]{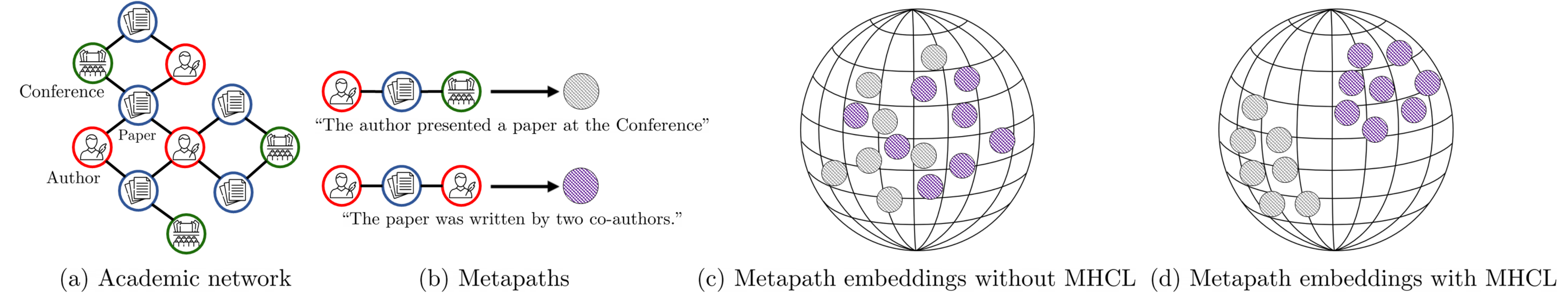}
    \caption{Basic concept of \algname{}.(Here, the striped circles represent the embedding of each metapath.)}
    \label{fig:Basic_concept}
\end{figure*}

Based on these observations, we propose a new framework Metapath-based Hyperbolic Contrastive Learning (\algname{}) for heterogeneous graph representation learning to effectively capture various semantic structural properties that inherently have power-law structures. Specifically, we design the intra-hyperbolic space attention mechanism to learn metapath instance representations and aggregate them in metapath-specific hyperbolic spaces to obtain metapath embeddings. The distinct curvature of metapath-specific hyperbolic spaces is set as a learnable parameter to find the appropriate curvature for each metapath during \algname{} training. 

Subsequently, in the unified hyperbolic space, the inter-hyperbolic space attention mechanism is used to aggregate semantic information from distinct metapaths, weighting their importance. When metapath embeddings are mapped into the unified hyperbolic space, embeddings with different semantic information may be represented close to each other and this distortion causes the \algname{} to mistakenly treat metapath embeddings with different semantic information as similar during the training process. To prevent this distortion and improve the discriminability of metapath embeddings, we propose a metapath-based hyperbolic contrastive learning method.

%To clearly distinguish metapath embeddings that represent different semantic and structural information within the unified hyperbolic space, metapath-based hyperbolic contrastive learning is adopted. 

Figure~\ref{fig:Basic_concept} provides examples of how metapath embeddings are distributed in the unified hyperbolic space, comparing cases with and without the proposed \algname{} method. In Figure~\ref{fig:Basic_concept}(c), without \algname{}, embeddings of different metapaths (represented in different colors) are mixed together without clear separation. This not only makes it difficult to distinguish heterogeneous information derived from diverse metapaths but also hinders effective use of the advantages from hyperbolic space. In contrast, as shown in Figure~\ref{fig:Basic_concept}(d), with \algname{} using our own contrastive learning, embeddings from the same metapaths form tighter clusters, while embeddings from different metapaths are more distinctly separated. This contrastive learning approach improves the discriminability of metapath embeddings, leading to more effective learning of heterogeneous information within heterogeneous graph and enabling us to use the advantages of hyperbolic space.

The main contributions of our work can be summarized as follows:
\begin{itemize}
    \item We propose a novel \algname{} framework for effective heterogeneous graph representation learning.
    \item We design attention mechanisms in multiple hyperbolic space to enhance representations of heterogeneous graphs. With multiple hyperbolic spaces, \algname{} effectively captures the diverse degree distributions corresponding to distinct metapaths.
    \item We design a hyperbolic contrastive learning method that learns relation between different metapaths and enhances the discriminability of heterogeneous information derived from distinct metapaths.
    \item We conduct comprehensive experiments to evaluate the effectiveness of \algname{}. The experimental results demonstrate that \algname{} outperforms state-of-the-art baselines in various downstream tasks with heterogeneous graphs.
\end{itemize}

The remainder of this paper is organized as follows. In Section~\ref{sec:related_work}, we reviews relevant works. In Section~\ref{sec:Preliminaries}, we introduce the preliminaries and essential background for our method. In Section~\ref{sec:Methodology}, we describe the details of the our proposed method \algname{}. In Section~\ref{sec:Experiments and Discussion}, we present experimental results and  analyses. Finally, in Section~\ref{sec:Conclusion}, we conclude this paper.

\section{Related work}
Our proposed method is related to three broader areas of research, namely Euclidean heterogeneous GNNs, hyperbolic heterogeneous GNNs, and hyperbolic graph contrastive learning.
\label{sec:related_work}
\subsection{Euclidean Heterogeneous GNNs}
\label{Eu:HGNN}
GNNs~\cite{wu2020comprehensive, zhou2020graph, khemani2024review} are effective in graph representation learning and there has been considerable interest in extending GNNs to heterogeneous graphs by considering the heterogeneity inherent in such graphs. Most recent studies~\cite{wang2022survey, bing2023heterogeneous} leverage metapaths or utilize link-type dependent graph convolution operations (e.g., link-type dependent attention mechanism) to capture semantic information and complex structures arising from heterogeneity.

HAN~\cite{wang2019heterogeneous} proposed a hierarchical graph attention networks model to aggregate information from different hierarchical levels: node-level attention and semantic-level attention. At the node-level attention, HAN aggregates information from metapath-based neighbors, while at the semantic level, it aggregates information from different metapaths. In node-level attention, intermediate nodes within metapath instances are ignored. MAGNN~\cite{fu2020magnn} extends HAN by considering intermediate nodes in metapath instances. Both HAN and MAGNN rely on predefined metapaths to capture rich semantic information within heterogeneous graphs. However, defining metapaths without domain-specific knowledge can be challenging across various datasets.

In order to address this challenge, many studies have proposed automatic metapath selection. GTN~\cite{yun2019graph} proposed graph transformer layers to learn a soft selection of link types and composite relations to generate useful metapaths. GraphMSE~\cite{li2021graphmse} proposed automatic metapath selection by leveraging semantic feature space alignment. Specifically, GraphMSE employs MLPs~\cite{hornik1989multilayer} to learn metapath instance representations and aligns these representations to facilitate automatic metapath selection.

On the contrary, there are heterogeneous GNNs that do not rely on metapath construction and instead utilize link-type dependent graph convolution operations to implicitly learn semantic information within heterogeneous graphs. HGT~\cite{hu2020heterogeneous} proposed node and link type-dependent attention mechanisms to capture the various types of relations within heterogeneous graphs. Simple-HGN~\cite{lv2021we} enhances graph attention mechanisms by incorporating link-type information into the calculation of attention scores.

\subsection{Hyperbolic Heterogeneous GNNs}
Due to the fact that Euclidean hetegeneous GNNs cannot effectively capture the structural properties inherent in heterogeneous graphs, such as hierarchical or power-law structures, recent heterogeneous GNNs that utilize the hyperbolic space as the embedding space have been proposed. One notable model is SHAN~\cite{li2023multi}, which discovered that the degree distributions of nodes in the simplicial complex extracted from heterogeneous graphs follow a power-law distribution. It utilizes graph attention mechanisms in hyperbolic space to learn such complex structures. While SHAN used hyperbolic space to learn the power-law structures of the simplicial complex, HHGAT~\cite{park2024hyperbolic} utilizes hyperbolic space to learn power-law structures based on metapaths to learn semantic structural information within heterogeneous graphs. 

Although these models have achieved remarkable performance, relying solely on one hyperbolic space presents limitations given the various power-law distributions within a heterogeneous graph. This is because a single hyperbolic space is unsuitable for representing and integrating power-law structures of different distributions. To address the limitation of a single hyperbolic space, McH-HGCN~\cite{liu2023mch} proposed hyperbolic heterogeneous GNNs with distinct curvatures corresponding to each link type and MSGAT~\cite{park2024multi} proposed hyperbolic heterogeneous GNNs that use multiple hyperbolic spaces to effectively learn diverse power-law structures corresponding to specific metapaths.

\subsection{Hyperbolic Graph Contrastive Learning}
Recently, graph contrastive learning~\cite{ju2024towards} has gained a lot of attention as an effective method to learn graph representations. Consequently, several studies have proposed hyperbolic graph contrastive learning which used hyperbolic distance as a similarity measure in hyperbolic space for contrastive learning. 
HGCL~\cite{liu2022enhancing} used two hyperbolic GNNs to generate different hyperbolic views and proposed hyperbolic contrastive loss to refine graph representations.
HCTS~\cite{yang2024hyperbolic} proposed the hyperbolic graph contrastive learning framework for cross-domain recommendation. Specifically, HCTS proposed three contrastive strategies: 1) user-user, 2) user-item, 3) item-item contrastive learning tasks. By incorporating these three aspects of contrastive learning, HCTS effectively transfers knowledge across domains from different perspectives.
HyperCL~\cite{Qin2024HGCL} performs hyperbolic contrastive learning using node representations obtained from both hyperbolic space and tangent space to leverage the geometric advantages of both spaces.

Although existing studies have been introduced and shows significant achievements, they have fundamental limitations. Prior Euclidean heterogeneous GNNs fail to effectively capture complex structures within heterogeneous graphs. While prior hyperbolic heterogeneous GNNs either fail to learn diverse power-law structures or lack the ability to distinguish among different semantic information from distinct metapaths.

In this work, unlike prior studies, \algname{} aim to learn diverse power-law structures within heterogeneous graphs and leverage geometric similarity or dissimilarity between different metapaths by utilizing metapath-based hyperbolic contrastive learning method which takes advantage of the hyperbolic space and improving separability of metapath embeddings with distinct semantic information.

\section{Preliminaries}
\label{sec:Preliminaries}
In this section, we present key concepts related to heterogeneous graphs and hyperbolic space as adopted in this work. Table~\ref{tab:notation} summarizes the notation that used in this paper. These notations will be formally defined in the following sections when we introduce our methodology and the technical details.

\subsection{Heterogeneous Graph}
\begin{definition}[\bf Heterogeneous graph]
A heterogeneous graph is defined as a graph $\mathcal{G} = (\mathcal{V}, \mathcal{E}, f_v(\cdot), f_e(\cdot))$. $\mathcal{V}$ is a set of nodes, $\mathcal{E}$ is a set of links, $f_v(\cdot) :\;\mathcal{V} \rightarrow T_\mathcal{V}$ is a node type mapping function, and $f_e(\cdot) :\;\mathcal{E} \rightarrow T_\mathcal{E}$ is a link type mapping function, where $T_\mathcal{V}$ and $\mathcal{T}_\mathcal{E}$ are sets of node types and link types, respectively, with $|T_\mathcal{V}| + |T_\mathcal{E}| > 2$.
\end{definition}

\begin{definition}[\bf Metapath]
A metapath $\phi$ is defined as a sequence of node types and links $A_1\xrightarrow{l_1}A_2\xrightarrow{l_2}\cdots\xrightarrow{l_i}A_{i+1}$ (abbreviated as $A_1,A_2,\cdots,A_{i+1}$); it can be expressed as a composite relation $l=l_1\circ l_2\circ\cdots\circ l_i$ between nodes $A_1$ and $A_{i+1}$, where $\circ$ denotes the composition operator on relations.
\end{definition}

\begin{definition}[\bf Metapath instance]
Given a metapath $\phi$ within the heterogeneous graph, a metapath instance $p$ of $\phi$ is defined as a sequence of nodes in the heterogeneous graph following the schema defined by $\phi$.
\end{definition}

\begin{table}[t!]
\centering
\caption{Notations used in this paper.}
\label{tab:notation}
\resizebox{\columnwidth}{!}{%
\begin{tabular}{cl}
\hline
\multicolumn{1}{l}{Notation} & Explanation \\ \hline
$\mathcal{G}$ & A Heterogeneous graph \\
$\Phi$ & The set of metapaths \\
$\mathcal{P}_v$ & The set of metapath instances starting from node $v$ \\
$\mathcal{V}_t$ & The set of embedding target nodes \\
$\mathbb{D}^{n,c}$ & $n$-dimensional \Poincare ball with curvature $-c\;(c>0)$ \\
$\mathbb{R}^n$ & $n$-dimensional Euclidean space \\
$\mathcal{T}_x\mathbb{D}^{n,c}$ & Tangent space at point $x$ \\
$exp^c_x(\cdot)$ & Exponential map at point $x$, $exp^c_x:\mathcal{T}_x\mathbb{D}^{n,c}\rightarrow\mathbb{D}^{n,c}$ \\
$log^c_x(\cdot)$ & Logarithmic map at point $x$, $log^c_x:\mathbb{D}^{n,c}\rightarrow \mathcal{T}_x\mathbb{D}^{n,c}$ \\
$\oplus_c$ & M\"obius addition \\
$\otimes_c$ & Hyperbolic matrix-vector multiplication \\
$\sigma\otimes_c(\cdot)$ & Hyperbolic non-linear activation function \\\hline
\end{tabular}%
}
\end{table}

\subsection{Hyperbolic Space}
The hyperbolic space has a constant negative curvature and expands exponentially. Due to these properties, the hyperbolic space is appropriate for modeling tree-like or hierarchically structured graphs where the number of nodes grows exponentially. Unlike Euclidean space, the hyperbolic space is not a vector space, and thus vector operations cannot be applied. Therefore, Definitions 4-8 are required to perform graph convolution operations in the hyperbolic space.

\begin{definition}[\bf \Poincare ball model]
\label{def:4}
\Poincare ball model with curvature $-c\ (c>0)$ is defined by the Riemannian manifold $(\mathbb{D}^{n,c}, g_x^c)$, where
\begin{align*}
    & \mathbb{D}^{n,c}=\{x \in \mathbb{R}^n : c||x||^2 < 1\},\\
    & g_x^c = \left(\lambda^c_x\right)I_d.
\end{align*}
Here, $\mathbb{D}^{n,c}$ is the open $n$-dimensional unit ball with radius $\frac{1}{\sqrt{c}}$ and $g_x^c$ is the Riemannian metric tensor, where $\lambda_x^c=\frac{2}{1-c||x||^2}$ and $I_d$ is the identity matrix. We denote $\mathcal{T}_x\mathbb{D}^{n,c}$ as the tangent space centered at point $x$.
\end{definition}

\begin{definition}[\bf \Mobius addition]
\label{def:5}Given the point $x,y \in \mathbb{D}^{n,c}$, \Mobius addition which represents the equation for the addition operation in the \Poincare ball model with curvature $-c\;(c>0)$ is defined as follows:
\begin{align*}
    x\oplus_c y = \frac{\left(1+2c\langle x,y\rangle+c||y||^2\right)x+\left(1-c||x||^2\right)y}{1+2c\langle x,y \rangle+c^2||x||^2||y||^2},
\end{align*}
where $\langle\cdot\rangle$ is the Euclidean inner product and $||\cdot||$ is the Euclidean norm.
\end{definition}

\begin{definition}[\bf Exponential and logarithmic maps]\label{def:6}
In \Poincare ball model with curvature $-c\;\left(c>0\right)$, the exponential map $exp^c_x:\mathcal{T}_x\mathbb{D}^{n,c}\rightarrow\mathbb{D}^{n,c}$ and the logarithmic map $log^c_x:\mathbb{D}^{n,c}\rightarrow \mathcal{T}_x\mathbb{D}^{n,c}$ are defined as shown below:
\begin{align*}
    & exp_x^c\left(v\right) = x\oplus_c\left(\text{tanh}\left(\sqrt{c}\frac{\lambda_x^c||v||}{2}\right)\frac{v}{\sqrt{c}||v||}\right),\\
    &log_x^c\left(y\right)=\frac{2}{\sqrt{c}\lambda_x^c}\text{tanh}^{-1}\left(\sqrt{c}||-x\oplus_cv||\right)\frac{-x\oplus_c y}{||-x\oplus_c y||},
\end{align*}
where $x$ and $y$ are the points in 
the hyperbolic space $\mathbb{D}^{n,c}$ and $x\neq y$. Here, $v$ is a nonzero tangent vector in the tangent space $\mathcal{T}_x\mathbb{D}^{n,c}$.
\end{definition}

\begin{definition}[\bf Hyperbolic matrix-vector multiplication] \label{def:7}
Given a point $x\in\mathbb{D}^{n,c}$ and a matrix $M\in\mathbb{R}^{m\times n}$, the matrix-vector multiplication operation in hyperbolic space is defined as follows:
\begin{align*}
    & M\otimes_c x = exp_0^c\left(Mlog_0^c\left(x\right)\right),
\end{align*}
where $\mathbf{0}\in\mathbb{R}^{n}$ is a zero vector. 
\end{definition}

\begin{definition}[\bf Hyperbolic non-linear activation function]\label{def:8}
Given the point $x\in\mathbb{D}^{n,c}$, the hyperbolic non-linear activation function is defined as follows:
\begin{align*}
    & \sigma\otimes_c\left(x\right) = exp_0^c\left(\sigma\left(log_0^c\left(x\right)\right)\right),
\end{align*}
where $\sigma$ is the Euclidean non-linear activation function.
\end{definition}

\begin{figure*}[t!]
\centering
    \includegraphics[width=\linewidth]{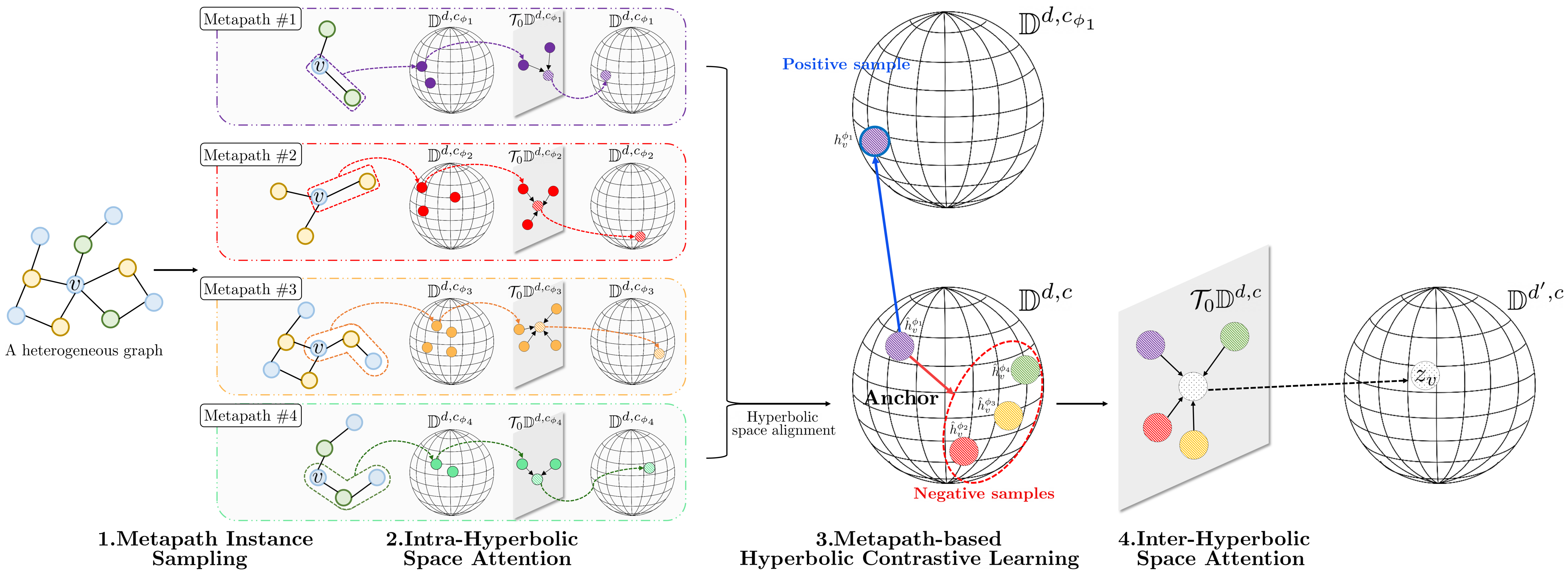}
    \caption{The framework of proposed \algname{}. Here, node $v$ indicates an embedding target node.}
    \label{figure:framework}
\end{figure*}

\section{Methodology}
\label{sec:Methodology}
In this section, we provide an introduction to our proposed model~\algname{}.

\subsection{Overview}
Figure~\ref{figure:framework} illustrates the overall procedure of \algname. Our proposed model consists of four main steps.
\begin{enumerate}
    \item First, all the metapath instances within a maximum length of $l$ are sampled starting from the embedding target nodes.
    \item Second, each set of metapath instances is embedded into its corresponding metapath-specific hyperbolic space to capture structural and semantic information. Metapath instance embeddings are then aggregated to obtain metapath embeddings.
    \item Third, a metapath-based hyperbolic contrastive learning method is applied to enhance the discriminability of the metapath embeddings.
    \item Fourth, metapath embeddings are aggregated to obtain node embeddings as outputs.
\end{enumerate}

\subsection{Metapath Instance Sampling}
First, to leverage the semantic structural properties within a heterogeneous graph $\mathcal{G}$, we sample a metapath instance set $\mathcal{P}_v$ for a given embedding target node $v$ in $\mathcal{V}_t$. Every metapath instance $p$ in $\mathcal{P}_v$ starts from embedding target node $v$ and has a length within a maximum metapath length $l$. For metapath instance sampling, we use the breadth-first search algorithm. This metapath instance sampling process enables us to handle the difficulties of defining metapaths based on domain-specific knowledge.

\subsection{Intra-Hyperbolic Space Attention}
\subsubsection{Hyperbolic mean-linear encoding} After metapath instance sampling, we construct a metapath instance encoder to transform all node features within a metapath instance into a single representation for each metapath instance. Specifically, we average the node features in Euclidean space and map the resulting single representation into a metapath-specific hyperbolic space. This transformation process can be formulated as follows:
\begin{align}
     x_p^{\mathbb{E}} &= \frac{1}{j}\sum_{i=1}^j x_i \left(j\leq l\right),\\
     x_p^{\mathbb{H,\phi}} &= W_t\otimes_{c_{\phi}} exp_0^{c_{\phi}}\left(x_p^{\mathbb{E}}\right).
\end{align}

In (1), $x_i \in \mathbb{R}^n$ denotes the features of node $i$ and $j$ denotes the length of the metapath instance $p$. In (2), given the Euclidean metapath instance feature $x_p^{\mathbb{E}}$, we first map $x_p^{\mathbb{E}}$ to a metapath-specific hyperbolic space $\mathbb{D}^{n,c_{\phi}}$ via the exponential map $exp_0^{c_{\phi}}:\mathcal{T}_0\mathbb{D}^{n,c_{\phi}}\rightarrow\mathbb{D}^{n,c_{\phi}}$. To adopt the exponential map, we assume $x_p^{\mathbb{E}}$ is included in the tangent space $\mathcal{T}_0\mathbb{D}^{n,c_{\phi}}$ at point $x=0$.

Note that $W_t\in\mathbb{R}^{n\times n}$ denotes a transformation matrix, $x_p^\mathbb{E} \in \mathbb{R}^{n}$ denotes the Euclidean feature of the metapath instance $p$ and $ x_p^{\mathbb{H,\phi}}\in\mathbb{D}^{n,c_{\phi}}$ denotes hyperbolic metapath instance feature. Here, $\mathbb{D}^{n,c_{\phi}}$ is a metapath-specific hyperbolic space that effectively represents structural properties of metapath instances following a specific metapath $\phi$. 

Additionally, $-c\;\left(c>0\right)$ is a learnable parameter that represents the negative curvature of hyperbolic space, and each metapath-specific hyperbolic space for metapath $\phi$ has a distinct negative curvature.

\subsubsection{Hyperbolic metapath instance embedding}
We use hyperbolic linear transformation with a hyperbolic non-linear activation function to obtain metapath instance embedding in metapath-specific hyperbolic space. This process is formulated as
\begin{align}
\label{eq:3}
    h_p^{\mathbb{H},\phi} &= \sigma\otimes_{c_{\phi}}\left(W_1\otimes_{c_{\phi}} x_p^{\mathbb{H,\phi}}\right)\oplus_{c_{\phi}} exp_0^{c_{\phi}}\left(b_1\right),
\end{align}
where $h_p^{\mathbb{H},\phi}\in\mathbb{D}^{d,c_{\phi}}$ is a latent representation of metapath instance $p$ in metapath-specific hyperbolic space $\mathbb{D}^{d,c_{\phi}}$. Here, $d$ is the dimension of hyperbolic space for latent metapath instance representations. Additionally, $W_1\in\mathbb{R}^{d\times n}$ is a weight matrix and $b_1\in\mathbb{R}^d$ is a bias vector. Note that $\otimes_{c_{\phi}}$ and $\oplus_{c_{\phi}}$ denote hyperbolic matrix-vector multiplication and \Mobius addition, respectively.

\subsubsection{Intra-metapath specific hyperbolic space attention}
After obtaining different latent metapath instance representations $h_p^{\mathbb{H},\phi}$, we define attention mechanisms in the metapath-specific hyperbolic space to aggregate them. First, we calculate the importance of each metapath instance $\alpha_p$ as follows:
\begin{align}
\label{eq:4}
    e_p &= a^T\otimes_{c_{\phi}} log_0^{c_{\phi}}\left(h_p^{\mathbb{H},\phi}\right), \\
    \alpha_p &= \frac{\text{exp}\left(e_p\right)}{\sum_{q\in\mathcal{P}_v^{\phi}}\text{exp}\left(e_q\right)},
\end{align}
where $e_p$ denotes the importance of each metapath instance $p \in \mathcal{P}_v^{\phi}$, $log_0^{c_{\phi}}:\mathbb{D}^{d,c_{\phi}}\rightarrow\mathcal{T}_0\mathbb{D}^{d,c_{\phi}}$ denotes the logarithmic map function, $\mathcal{P}_v^{\phi}$ denotes a subset of $\mathcal{P}_v$ consisting of metapath instances that follow a specific metapath $\phi$, and $a\in\mathbb{R}^{d}$ is an attention vector for the metapath instance. After calculating the importance of each metapath instance, we adopt the softmax function to these values to obtain the weight of each metapath instance.

Then, the metapath embedding for node $v$ is obtained from the weight of each metapath instance and their latent representations in metapath-specific hyperbolic space. This process can be formulated as
\begin{align}
\label{eq:6}
    h_v^\phi &= \sigma \otimes_{c_{\phi}} \bigg(exp_0^{c_{\phi}} \bigg( \sum_{p \in \mathcal{P}_v^{\phi}} \alpha_p \cdot log_0^{c_{\phi}} \big( h_p^{\mathbb{H}, \phi} \big) \bigg) \bigg),
\end{align}
where $h_v^\phi\in\mathbb{D}^{d,c_{\phi}}$ denotes metapath embedding for node $v$. Note that in (\ref{eq:3}) and (\ref{eq:6}), $\sigma \otimes_{c_{\phi}}$ denotes the hyperbolic non-linear activation function with LeakyReLU.
\subsubsection{Hyperbolic multi-head attention}
We introduce multi-head attention in the hyperbolic space to enhance metapath embeddings and stabilize the learning process. Specifically, we divide the attention mechanisms into $K$ independent attention mechanisms, conduct them in parallel, and then concatenate the metapath embedding from each attention mechanism to obtain the final metapath embedding $h_v^\phi$, which can be formulated as
\begin{align}
    h_v^\phi = \parallel_{k=1}^{K} \sigma \otimes_{c_{\phi}} \bigg( exp_0^{c_{\phi}} \bigg( \sum_{p \in \mathcal{P}_v^{\phi}} \alpha_p^k \cdot log_0^{c_{\phi}} \big(h_{p}^{\mathbb{H}, \phi} \big) \bigg) \bigg).
\end{align}

\begin{figure}[t!]
\centering
    \includegraphics[width=0.9\columnwidth]{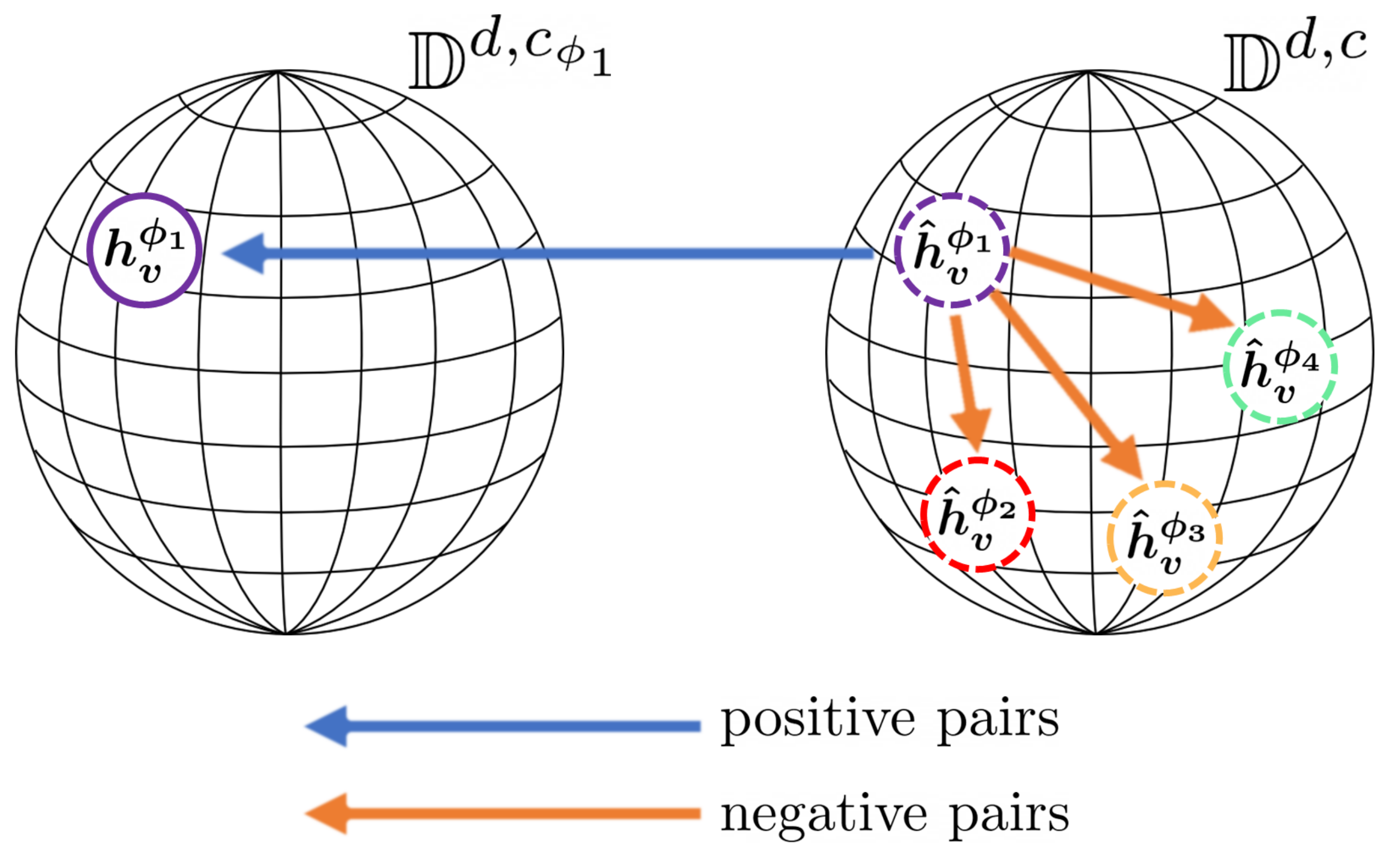}
    \caption{Example of the sample selection strategy.}
    \label{figure:sample selection}
\end{figure}

\subsection{Metapath-based Hyperbolic Contrastive Learning}
After obtaining the metapath embeddings in metapath-specific hyperbolic spaces for each metapath, we adopt a metapath-based hyperbolic contrastive learning method to enhance these representations, which was not presented in our earlier studies~\cite{park2024hyperbolic, park2024multi}.
\subsubsection{Hyperbolic space alignment}
To perform metapath-based hyperbolic contrastive learning between metapath embeddings included in metapath-specific hyperbolic spaces with different negative curvatures, we have to map the metapath embeddings into the unified hyperbolic space for integration, which is non-trivial and challenging. Specifically, since direct operations between metapath embeddings included in different metapath-specific hyperbolic spaces are not applicable, it is essential to first align them within the unified hyperbolic space. 
For hyperbolic space alignment, we leverage the property that metpath-specific hyperbolic spaces with different negative curvatures share a common tangent space at the north pole. This process can be formulated as
\begin{align}
\label{eq:8}
    \hat{h}_v^{\phi} = exp_0^{c}(W_2\cdot log_0^{c_{\phi}}(h_v^\phi)).
\end{align}

As shown in (\ref{eq:8}), we first map metapath embeddings $h_v^\phi$ into a tangent space of metapath-specific hyperbolic space via the logarithmic map function  $log_0^{c_{\phi}}:\mathbb{D}^{d,c_{\phi}}\rightarrow\mathcal{T}_0\mathbb{D}^{d,c_{\phi}}$. Next, using the transformation matrix $W_2\in\mathbb{R}^{d\times d}$, we map the metapath embeddings from $\mathcal{T}_0\mathbb{D}^{d,c_{\phi}}$ to $\mathcal{T}_0\mathbb{D}^{d,c}$. Here, $\mathcal{T}_0\mathbb{D}^{d,c_{\phi}}$ and $\mathcal{T}_0\mathbb{D}^{d,c}$ denote the tangent space of metapath-specific hyperbolic spaces and the unified hyperbolic space, respectively.
Finally, using the exponential map $exp_0^{c}:\mathcal{T}_0\mathbb{D}^{d,c}\rightarrow\mathbb{D}^{d,c}$, we map the metapath embeddings to the unified hyperbolic space $\mathbb{D}^{d,c}$. Note that $\hat{h}_v^{\phi}$ denotes the metapath embedding within the unified hyperbolic space.

\subsubsection{Sample selection strategy}
\label{sec:sample selection}
To perform metapath-based hyperbolic contrastive learning, we select postive and negative samples based on their related metapaths. Figure~\ref{figure:sample selection} illustrates an example of the sample selection strategy of \algname{}. 

Given a metapath embedding $\hat{h}_v^{\phi} \in \mathbb{D}^{d,c}$, we select $h_v^{\phi} \in \mathbb{D}^{d,c_{\phi}}$ as its positive sample to preserve the consistency of the same metapath across different hyperbolic spaces during training. For negative samples, we select $\hat{h}_v^{j(\phi)} \in \mathbb{D}^{d,c}$, where $j(\phi)\in\Phi\setminus\{\phi\}$, to maximize the discriminability from metapaths with different semantics. Specifically, with the selected negative samples, \algname{} is trained to distinguish between metapath embeddings that are structurally similar but semantically different.

\subsubsection{Hyperbolic contrastive learning}
Based on the hyperbolic space alignment and sample selection strategy, we perform hyperbolic contrastive learning. Given the metapath embedding $\hat{h}_v^{\phi}$ within the unified hyperbolic space, we first project the corresponding positive sample $h_v^{\phi}$ from its metapath-specific hyperbolic space to the tangent space via the logarithmic map function $log_0^{c_{\phi}}:\mathbb{D}^{d,c_{\phi}}\rightarrow\mathcal{T}_0\mathbb{D}^{d,c_{\phi}}$.
Then, we map $log_0^{c_{\phi}}(h_v^\phi)$ into the unified hyperbolic space using the exponential map function $exp_0^c : \mathcal{T}_0 \mathbb{D}^{d,c_{\phi}} \rightarrow \mathbb{D}^{d,c}$. This transformation enables the computation of similarity between $\hat{h}_v^{\phi}$ and $h_v^{\phi}$ within the unified hyperbolic space. This positive sample transformation process can be formulated as
\begin{align}
    p(h_v^{\phi}) = exp_0^{c}(log_0^{c_{\phi}}(h_v^\phi)).
\end{align}

Then, with positive and negative samples, we can finally formulate metapath-based hyperbolic contrastive learning as
\begin{align}
    \mathcal{L}_{hyp} &= \sum_{\phi\in\Phi}\frac{\text{exp}(-d_c(\hat{h}_v^{\phi}, p(h_v^{\phi})/\tau))}{\sum_{j(\phi)\in\Phi\setminus\{\phi\}}\text{exp}(-d_c(\hat{h}_v^{\phi}, \hat{h}_v^{j(\phi)})/\tau)},\\
    d_c(x,y) &= \frac{1}{\sqrt{|c|}}\text{cosh}^{-1}\bigg(1-\frac{2c||x-y||^2_2}{(1+c||x||^2_2)(1+c||y||^2_2)}\bigg).
\end{align}

In (10), given the metapath embedding $\hat{h}_v^{\phi}$, $p(h_v^{\phi})$ and $\hat{h}_v^{j(\phi)}$ denote positive and negative samples of $\hat{h}_v^{\phi}$, respectively. In (11), $d_c(x,y)$ represents hyperbolic distance between two points $x$ and $y$ in the \Poincare ball model with curvature $-c (c>0)$. By using the distance in the hyperbolic space as a metric of similarity between two samples, we can leverage the advantages of the hyperbolic space, which grows exponentially due to its negative curvature.

\begin{algorithm}[t!]
\caption{\small{\algname{} Training}}
\label{alg:process}
\begin{algorithmic}
\begin{small}
\REQUIRE Embedding target node $v$, The set of metapaths $\Phi$, The set $\mathcal{P}_v$ of metapath instances starting from node $v$, The set of node features $X$, The number of attention heads $K$, hyperparameter $\lambda$.
%\ENSURE Prediction for downstream tasks.
\end{small}
\end{algorithmic}
\begin{algorithmic}[1]
\begin{small}
\setstretch{1.2}
\FOR{$\phi\in\Phi$}
\FOR{$k=1\cdots K$}
\FOR{$p\in\mathcal{P}_v^\phi\;(\mathcal{P}_v^\phi\subset \mathcal{P}_v)$}
\STATE{\textbf{(1) Intra-Hyperbolic Space Attention}}
\STATE{$x_p^{\mathbb{E}}\leftarrow\frac{1}{|p|}\sum_{i=1}^{|p|} x_i\;(\forall x_i \in X);$}
\STATE{$x_p^{\mathbb{H,\phi}}\leftarrow W_t\otimes_{c_{\phi}} exp_0^{c_{\phi}}(x_p^{\mathbb{E}});$}
\STATE{$ h_p^{\mathbb{H},\phi}\leftarrow\sigma\otimes_{c_{\phi}}(W_1\otimes_{c_{\phi}} x_p^{\mathbb{H,\phi}})\oplus_{c_{\phi}} exp_0^{c_{\phi}}(b_1);$}
\STATE{$\alpha_p = \text{softmax}(a^T\otimes_{c_{\phi}} log_0^{c_{\phi}}(h_p^{\mathbb{H},\phi}));$}
\ENDFOR
\STATE{$h_v^\phi\leftarrow\sigma\otimes_{c_{\phi}}(exp_0^{c_{\phi}}(\sum_{p\in\mathcal{P}_v^{\phi}}\alpha_p\cdot log_0^{c_{\phi}}(h_p^{\mathbb{H}, \phi})));$}
\ENDFOR
\STATE{$h_v^\phi\leftarrow\parallel_{k=1}^{K}\sigma\otimes_{c_{\phi}}(exp_0^{c_{\phi}}(\sum_{p\in\mathcal{P}_v^{\phi}}\alpha_p^k\cdot log_0^{c_{\phi}}(h_{p}^{\mathbb{H}, \phi})));$}
\STATE{\textbf{(2) Metapath-based Hyperbolic Contrastive Learning}}
\STATE{$\hat{h}_v^{\phi}\leftarrow exp_0^{c}(W_2\cdot log_0^{c_{\phi}}(h_v^\phi));$}
\STATE{$p(h_v^{\phi})\leftarrow exp_0^{c}(log_0^{c_{\phi}}(h_v^\phi));$}
\STATE{Calculate $\mathcal{L}_{hyp}$ in (10);}
\STATE{\textbf{(3) Inter-Hyperbolic Space Attention}}
\STATE{$\beta_{\phi}\leftarrow \text{softmax}(\sigma\otimes_c((W_3\otimes_c\hat{h}_v^{\phi})\oplus_c exp_0^c(b_3))));$}
\ENDFOR
\STATE{$z_v\leftarrow exp_0^c(\sum_{\phi\in\Phi}\beta_\phi\cdot log_o^c(\hat{h}_v^{\phi}));$}
\STATE{$f(z_v)\leftarrow\sigma(W_o\cdot(log_0^c(z_v)));$}
\STATE{Calculate $\mathcal{L}_{task}$ in (16) or (17);}
\STATE{Minimize $\mathcal{L}_{task}+\lambda\mathcal{L}_{hyp}$;}
%\RETURN{Obtain prediction for downstream tasks;}
\end{small}
\end{algorithmic}
\end{algorithm}

\subsection{Inter-Hyperbolic Space Attention}
Once metapath embeddings $\hat{h}_v^{\phi}\in\mathbb{D}^{d,c}$ are obtained for each metapath $\phi\in\Phi$, we aggregate them using the proposed inter-hyperbolic space attention mechanism, which learns the importance of each metapath embedding and assigns higher weights to those that are semantically more important. First, we calculate the importance of each metapath embeddings $\beta_{\phi}$ as follows:
\begin{align}
    e_{\phi} &= b^{T}\otimes_c(\sigma\otimes_c((W_3\otimes_c\hat{h}_v^{\phi})\oplus_c exp_0^c(b_3))),\\
    \beta_{\phi} &= \frac{\text{exp}(e_{\phi})}{\sum_{\pi\in\Phi}\text{exp}(e_{\pi})}.
\end{align}
where $e_\phi$ denotes the importance of each metapath embedding. After calculating $e_\phi$, we normalize these values using the softmax function to obtain their weights. Note that $W_3\in\mathbb{R}^{d'\times d}$ denotes a weight matrix for the metapath embeddings, $b_3\in\mathbb{R}^{d'}$ is a bias vector, and $b\in\mathbb{R}^{d'}$ is an attention vector.  Then, the representation of node $v$, $z_v\in\mathbb{D}^{d',c}$ is calculated as a weighted sum:
\begin{align}
    z_v = exp_0^c(\sum_{\phi\in\Phi}\beta_\phi\cdot log_o^c(\hat{h}_v^{\phi})).
\end{align}

\subsection{Model Training}
We use the following non-linear transformation $f(\cdot)$ to map node representation $z_v$ into a Euclidean space with the desired output dimension, conducting various downstream tasks:
\begin{align}
\label{eq:12}
    f(z_v) = \sigma(W_o\cdot(log_0^c(z_v))),
\end{align}
where $W_o\in\mathbb{R}^{d_o\times d'}$ denotes the weight matrix, $d_o$ denotes the dimension of output vector, and $\sigma$ is the activation function.
We train \algname{} by minimizing the task loss $\mathcal{L}_{task}$ and the hyperbolic contrastive loss $\mathcal{L}_{hyp}$.
For node-level tasks is expressed as
\begin{align}
    \mathcal{L}_{task} &=-\sum_{v\in V_t}\sum_{c=1}^Cy_v[c]\cdot\text{log}\left(f(z_v)[c]\right),
\end{align}
where $v_t$ is the target node set extracted from the labeled node set, $C$ is the number of classes, $y_v$ is the one-hot encoded label vector for node $v$, and $f(z_v)$ is a vector predicting the label probabilities of node $v$.

For link-level task, $\mathcal{L}_{task}$ is expressed as
\begin{align}
\begin{split}
    \mathcal{L}_{task}  = &-\frac{1}{|\mathcal{S}|}\sum_{\left(u,v,y\right)\in\mathcal{S}}y\cdot \text{log}\left(\sigma\left(f(z_u)^Tf(z_v)\right)\right)\\&+\left(1-y\right)\text{log}\left(1-\sigma\left(f(z_u)^Tf(z_v)\right)\right),
\end{split}
\end{align}
where $\mathcal{S}$ is the set that includes both positive and negative node pairs, $y$ is the ground truth label for a node pair $(u, v)$. $\sigma$ is the sigmoid function, $z_u$ and $z_v$ are the embedding vectors of nodes $u$ and $v$, respectively.

The whole loss function for training \algname{} is then expressed as
\begin{align}
    \mathcal{L} &= \mathcal{L}_{task}+\lambda\mathcal{L}_{hyp},
\end{align}
where $\lambda$ is hyperparameters ranging from 0 to 1.

\subsection{Time Complexity Analysis}
\label{sec:complexity}
\algname{} consists of three main components for heterogeneous graph embedding: metapath instance sampling, intra-hyperbolic space attention, and inter-hyperbolic space attention. First, for metapath instance sampling from a heterogeneous graph $\mathcal{G}=(\mathcal{V}, \mathcal{E})$, we employ breadth-first search. Therefore, the time complexity of metapath instance sampling is $\mathcal{O}(|\mathcal{V}|+|\mathcal{E}|)$. Second, the time complexity of intra-hyperbolic space attention is $\mathcal{O}(|\mathcal{P}|nd)$, where $|\mathcal{P}|$ denotes the number of sampled metapath instances, $n$ and $d$ denote the dimension of initial node features and metapath embedding, respectively. Finally, the time complexity of inter-hyperbolic space attention is $\mathcal{O}(|\Phi|dd')$, where $|\Phi|$ denotes the number of metapaths and $d'$ denotes the dimension of node embedding. As a result, the overall time complexity of~\algname{} is $\mathcal{O}(|\mathcal{V}|+|\mathcal{E}|+|\mathcal{P}|nd+|\Phi|dd')$. Since $|\mathcal{P}|$ and $|\Phi|$ increase as the maximum metapath length $l$ increases, the overall time complexity of \algname{} increases linearly with respect to $l$.

\section{Experiments and Discussion}
\label{sec:Experiments and Discussion}
In this section, we analyze the efficiency of our proposed model~\algname{}, through with four real-world datasets and synthetic datasets. We compare~\algname{} with several state-of-the-art GNN models. Our experiments are designed to answer the following research questions (RQs).
\begin{itemize}
    \item\textbf{RQ1}: Does \algname{} outperform the state-of-the-art baselines in node-level downstream tasks (RQ1-1) and link-level downstream tasks (RQ1-2)? 
     \item\textbf{RQ2}: How does each main component of \algname{} contribute to the learning of heterogeneous graph representations?
     \item\textbf{RQ3}: Does \algname{} exhibit better robustness than the other baselines?
     \item\textbf{RQ4}: How do hyperparameters affect the performance of \algname{}? 
     \item\textbf{RQ5}: Does \algname{} obtains better metapath representations via metapath-based hyperbolic contrastive learning?
\end{itemize}

\begin{table}[t]
\centering
\caption{Statistics of real-world datasets}
\label{tab:datastats}
\resizebox{0.9\columnwidth}{!}{%
\begin{tabular}{ccccc} %\hline
\multicolumn{5}{c}{Node Classification and Clustering} \\\hline
Dataset & \# Nodes & \# Links & \# Classes & \# Features \\\hline
IMDB    & \begin{tabular}[c]{@{}c@{}}Movie (M) : 4,661\\ Director (D) : 2,270\\ Actor (A) : 5,841\end{tabular}   & \begin{tabular}[c]{@{}c@{}}M-D : 4,661\\ M-A : 13,983\end{tabular} & 3  & 1,256\\\hline
DBLP    & \begin{tabular}[c]{@{}c@{}}Author (A) : 4,057\\Paper (P) : 14,328\\ Conference (C) : 20\end{tabular}   & \begin{tabular}[c]{@{}c@{}}A-P : 19,645\\ P-C : 14,328\end{tabular} & 4  & 334\\\hline
ACM    & \begin{tabular}[c]{@{}c@{}}Paper (P) : 3,020\\ Author (A) : 5,912\\ Subject (S) : 57\end{tabular}   & \begin{tabular}[c]{@{}c@{}}P-A : 9,936\\ P-S : 3,025\\\end{tabular} & 3 & 1,902\\\hline
\\
\multicolumn{5}{c}{Link Prediction} \\\hline
Dataset & \# Nodes & \# Links & Target & \# Features \\\hline
LastFM    & \begin{tabular}[c]{@{}c@{}}User (U) : 1,892\\ 
Artist (A) : 17,632\\ Tag (T) : 1,088\end{tabular}   & \begin{tabular}[c]{@{}c@{}}U-A : 85,689\\ A-T : 21,553\end{tabular} & User-Artist  & 20,612\\\hline
\end{tabular}%
}
\end{table}

\subsection{Datasets}
To evaluate the performance of \algname{} on downstream tasks, we use four real-world heterogeneous graph datasets. Table~\ref{tab:datastats} shows the statistics of these datasets. Additionally, detailed descriptions of the datasets are provided below:
\begin{itemize}
    \item {\textbf{IMDB}\footnote{https://www.imdb.com/}} is an online database pertaining to movies and television programs. It comprises three types of nodes \{Movie (M), Director (D), Actor (A)\} and two types of links \{MD, MA\}. The movie type nodes are labeled into three classes based on the movie's genre \{Action, Drama, Comedy\}. The features of nodes are represented as bag-of-words of keywords.
    \item {\textbf{DBLP}\footnote{https://dblp.uni-trier.de/}} is a citation network that comprises three types of nodes \{Author (A), Paper (P), Conference (C)\} and two types of links \{AP, PC\}. The author type nodes are labeled into four classes based on the author's research area \{Database, Data Mining, Machine Learning, Information Retrieval\}. The features of nodes are represented as bag-of-words of keywords.
    \item {\textbf{ACM}\footnote{https://dl.acm.org/}} is a citation network that comprises three types of nodes \{Paper (P), Author (A), Subject (S)\} and two types of links \{PA, PS\}. The paper type nodes are labeled into three classes based on the paper's subject area \{Database, Wireless Communication, Data Mining\}. The features of nodes are represented as bag-of-words of plots.
    \item {\textbf{LastFM}\footnote{https://www.last.fm/}} is an online music website database. We extract a subset of LastFM with three types of nodes \{User (U), Artist (A), Tag (T)\} and two types of links \{UA, AT\}. This dataset is used for link prediction task between user and artist. The features of nodes are represented as one-hot vector representations.
\end{itemize}

\subsection{Competitors}
We compare \algname{} with several state-of-the-art GNNs categorized into four groups:
\begin{enumerate}[\bf i)]
  \item{\bf Euclidean homogeneous GNNs}: GCN and GAT.
  \item{\bf Hyperbolic homogeneous GNNs}: HGCN and HGCL.
  \item{\bf Euclidean heterogeneous GNNs}: HAN, MAGNN, GTN, HGT, and Simple-HGN.
  \item{\bf Hyperbolic heterogeneous GNNs}: McH-HGCN, SHAN, HHGAT, and MSGAT.
\end{enumerate}
For homogeneous GNNs, features are processed to be homogeneous for pair comparison with heterogeneous GNNs.

Details of the competitors are provided as follows:
\begin{itemize}
    \item {\textbf{GCN}}~\cite{kips2017iclr} performs convolution operations in the Fourier domain for homogeneous graphs. %We consider all node types as a single node type to perform on heterogeneous graphs.
    \item {\textbf{GAT}}~\cite{velickovic2018iclr} introduces convolution operations with attention mechanism for homogeneous graphs. %Similarly, we treat all node types as one when applying GAT on heterogeneous graphs.
    \item {\textbf{HGCN}}~\cite{chami2019hyperbolic} proposed graph neural neural networks that utilize the hyperbolic space as an embedding  to effectively learn hierarchical structures within homogeneous graphs.
    \item {\textbf{HGCL}}~\cite{liu2022enhancing} proposed hyperbolic graph contrastive learning method to enhance graph representations in the hyperbolic space.
    \item {\textbf{HAN}}~\cite{wang2019heterogeneous} proposes a graph attention network for heterogeneous graphs, which considers both node-level and semantic-level attention mechanisms.
    \item {\textbf{MAGNN}}~\cite{fu2020magnn} utilizes intra-metapath aggregation and inter-metapath aggregation to incorporate intermediate semantic nodes and multiple metapaths, respectively.
    \item {\textbf{GTN}}~\cite{yun2019graph} transforms heterogeneous graphs into multiple metapath graphs via Graph Transformer layers. The generated graphs facilitate more effective node representations.
    \item {\textbf{HGT}}~\cite{hu2020heterogeneous} introduces a heterogeneous sub-graph sampling algorithm to model Web-scale graph data, utilizing node and link-type dependent parameters to capture the heterogeneous over each link. 
    \item {\textbf{Simple-HGN}}~\cite{lv2021we} proposes link type aware graph attention mechanisms to handle the heterogeneity from multiple types of links that included in heterogeneous graphs.
    \item {\textbf{McH-HGCN}}~\cite{liu2023mch} introduces hyperbolic heterogeneous graph neural networks with distinct curvatures corresponding to each link type.
    \item {\textbf{SHAN}}~\cite{li2023multi} introduces hyperbolic heterogeneous graph attention networks to effectively learn multi-order relations that follows power-law distributions. Specifically, they extract simplical complexes from heterogeneous graphs and embed them in the hyperbolic space.
    \item {\textbf{HHGAT}}~\cite{park2024hyperbolic} proposed hyperbolic heterogeneous graph attention networks with a single hyperbolic space to learn power-law structures, effectively capturing such structural properties within heterogeneous graphs.
    \item {\textbf{MSGAT}}~\cite{park2024multi} proposed multi hyperbolic space-based heterogeneous graph attention networks to learn various power-law structures based on metapaths.
\end{itemize}

\begin{table*}[]
\centering
\caption{Experimental results (\%) for the node classification task. The best and second-best performers are bolded and underlined, respectively.}
\label{tab:node_classification}
\resizebox{\textwidth}{!}{%
\begin{tabular}{ccccccccccccccccc}
\hline
\multirow{2}{*}{Dataset} &
  \multirow{2}{*}{Metric} &
  \multirow{2}{*}{Train \%} &
  \multicolumn{2}{c}{\lowercase\expandafter{\romannumeral1}} & \multicolumn{2}{c}{\lowercase\expandafter{\romannumeral2}} & \multicolumn{5}{c}{\lowercase\expandafter{\romannumeral3}} & \multicolumn{5}{c}{\lowercase\expandafter{\romannumeral4}}\\ \cmidrule(lr){4-5} \cmidrule(lr){6-7} \cmidrule(lr){8-12} \cmidrule(lr){13-17}
  &
   &
   &
  GCN &
  GAT &
  HGCN &
  HGCL &
  HAN &
  MAGNN &
  GTN &
  HGT &
  Simple-HGN &
  McH-HGCN &
  SHAN &
  HHGAT &
  MSGAT &
  \algname{} \\ \hline
\multirow{8}{*}{IMDB} &
  \multirow{4}{*}{Macro-F1} &
  20\% &
  52.17\footnotesize$\pm$0.35 &
  53.68\footnotesize$\pm$0.26 &
  54.38\footnotesize$\pm$0.48 &
  56.86\footnotesize$\pm$0.71 &
  56.19\footnotesize$\pm$0.51 &
  59.33\footnotesize$\pm$0.38 &
  58.74\footnotesize$\pm$0.74 &
  56.14\footnotesize$\pm$0.65 &
  59.97\footnotesize$\pm$0.61 &
  58.16\footnotesize$\pm$0.49&
   62.23\footnotesize$\pm$0.76&
   63.16\footnotesize$\pm$0.39&
   \underline{65.75\footnotesize$\pm$0.81} &
   \bf66.11\footnotesize$\pm$0.73
   \\
 &
   &
  40\% &
  53.20\footnotesize$\pm$0.45 &
  56.33\footnotesize$\pm$0.71 &
  57.05\footnotesize$\pm$0.43 &
  58.44\footnotesize$\pm$0.65 &
  56.84\footnotesize$\pm$0.37 &
  60.70\footnotesize$\pm$0.48 &
  59.71\footnotesize$\pm$0.54 &
  57.12\footnotesize$\pm$0.53 &
  61.94\footnotesize$\pm$0.39 &
   60.31\footnotesize$\pm$0.56&
   63.98\footnotesize$\pm$0.68&
   65.07\footnotesize$\pm$0.63&
   \underline{68.07\footnotesize$\pm$0.54} &
   \bf69.08\footnotesize$\pm$0.68
   \\
 &
   &
  60\% &
  54.35\footnotesize$\pm$0.46 &
  56.93\footnotesize$\pm$0.54 &
   57.86\footnotesize$\pm$0.56&
   58.90\footnotesize$\pm$0.88 &
  58.95\footnotesize$\pm$0.71 &
  60.68\footnotesize$\pm$0.56 &
  61.88\footnotesize$\pm$0.50 &
  61.52\footnotesize$\pm$0.57 &
  66.73\footnotesize$\pm$0.42&
   61.93\footnotesize$\pm$0.33&
   66.68\footnotesize$\pm$0.71&
   65.72\footnotesize$\pm$0.56&
   \underline{71.42\footnotesize$\pm$0.48} &
   \bf72.29\footnotesize$\pm$0.59
   \\
 &
   &
  80\% &
  54.19\footnotesize$\pm$0.29 &
  57.25\footnotesize$\pm$0.18 &
  57.92\footnotesize$\pm$0.32 &
  59.65\footnotesize$\pm$0.74 &
  58.61\footnotesize$\pm$0.63 &
  61.15\footnotesize$\pm$0.55 &
  62.08\footnotesize$\pm$0.62 &
  63.69\footnotesize$\pm$0.59 &
  67.56\footnotesize$\pm$0.45 &
   62.29\footnotesize$\pm$0.50&
   68.49\footnotesize$\pm$0.67&
   67.42\footnotesize$\pm$0.51&
  \underline{70.03\footnotesize$\pm$0.60} &
   \bf71.27\footnotesize$\pm$0.91
   \\ \cline{2-17} 
 &
  \multirow{4}{*}{Micro-F1} &
  20\% &
  52.13\footnotesize$\pm$0.38 &
  53.67\footnotesize$\pm$0.31 &
  54.46\footnotesize$\pm$0.42&
  55.52\footnotesize$\pm$0.79 &
  56.71\footnotesize$\pm$0.53 &
  58.30\footnotesize$\pm$0.39 &
  61.97\footnotesize$\pm$0.63 &
  57.97\footnotesize$\pm$0.76 &
  63.76\footnotesize$\pm$0.60 &
   61.28\footnotesize$\pm$0.37&
   64.31\footnotesize$\pm$0.82&
   65.76\footnotesize$\pm$0.66&
   \underline{69.09\footnotesize$\pm$0.93}&
   \bf69.54\footnotesize$\pm$0.82
   \\
 &
   &
  40\% &
  53.34\footnotesize$\pm$0.41 &
  53.99\footnotesize$\pm$0.65 &
  57.02\footnotesize$\pm$0.46 &
  59.50\footnotesize$\pm$0.61 &
  56.68\footnotesize$\pm$0.70 &
  58.34\footnotesize$\pm$0.58 &
  62.10\footnotesize$\pm$0.53 &
  58.80\footnotesize$\pm$0.65 &
  65.60\footnotesize$\pm$0.46 &
   63.09\footnotesize$\pm$0.12&
   66.56\footnotesize$\pm$0.73&
   66.34\footnotesize$\pm$0.70&
   \underline{70.95\footnotesize$\pm$0.54}&
   \bf71.71\footnotesize$\pm$0.66
   \\
 &
   &
  60\% &
  54.61\footnotesize$\pm$0.42 &
  56.26\footnotesize$\pm$0.51 &
  58.01\footnotesize$\pm$0.50 &
  59.94\footnotesize$\pm$0.95 &
  58.26\footnotesize$\pm$0.82 &
  60.71\footnotesize$\pm$0.70 &
  63.55\footnotesize$\pm$0.39 &
  62.63\footnotesize$\pm$0.58 &
  69.29\footnotesize$\pm$0.74 &
   64.16\footnotesize$\pm$0.29&
   69.57\footnotesize$\pm$0.76&
   70.40\footnotesize$\pm$0.51&
   \underline{73.60\footnotesize$\pm$0.39}&
   \bf74.63\footnotesize$\pm$0.80
   \\
 &
   &
  80\% &
  54.37\footnotesize$\pm$0.33 &
  57.23\footnotesize$\pm$0.29 &
  58.54\footnotesize$\pm$0.93 &
  60.08\footnotesize$\pm$0.82 &
  59.35\footnotesize$\pm$0.65 &
  61.70\footnotesize$\pm$0.39 &
  65.57\footnotesize$\pm$0.91 &
  67.01\footnotesize$\pm$0.47 &
  69.35\footnotesize$\pm$0.66 &
   64.96\footnotesize$\pm$0.42&
  69.42\footnotesize$\pm$0.56 &
   69.61\footnotesize$\pm$0.89&
   73.37\footnotesize$\pm$0.59&
   \bf74.24\footnotesize$\pm$0.96
   \\\hline
\multirow{8}{*}{DBLP} &
  \multirow{4}{*}{Macro-F1} &
  20\% &
  87.51\footnotesize$\pm$0.15 &
  91.52\footnotesize$\pm$0.34 &
  91.69\footnotesize$\pm$0.38 &
  92.85\footnotesize$\pm$0.53 &
  92.63\footnotesize$\pm$0.46 &
  93.21\footnotesize$\pm$0.64 &
  92.45\footnotesize$\pm$0.37 &
  90.36\footnotesize$\pm$0.62 &
  93.48\footnotesize$\pm$0.56 &
  90.63\footnotesize$\pm$0.72&
  94.27\footnotesize$\pm$0.16&
   94.19\footnotesize$\pm$0.08&
  \underline{95.44\footnotesize$\pm$0.17}&
  \bf95.63\footnotesize$\pm$0.25
   \\
 &
   &
  40\% &
  88.55\footnotesize$\pm$0.46 &
  91.07\footnotesize$\pm$0.39 &
  91.93\footnotesize$\pm$0.35 &
  93.12\footnotesize$\pm$0.47 &
  92.35\footnotesize$\pm$0.64 &
  93.51\footnotesize$\pm$0.29 &
  92.39\footnotesize$\pm$0.41 &
  91.57\footnotesize$\pm$0.29 &
  93.98\footnotesize$\pm$0.27 &
   91.74\footnotesize$\pm$0.62&
  94.33\footnotesize$\pm$0.08&
  94.27\footnotesize$\pm$0.10&
  \underline{95.54\footnotesize$\pm$0.12}&
  \bf95.71\footnotesize$\pm$0.47
   \\
 &
   &
  60\% &
  89.44\footnotesize$\pm$0.27 &
  91.51\footnotesize$\pm$0.46 &
  92.60\footnotesize$\pm$0.89 &
  93.57\footnotesize$\pm$0.86 &
  92.86\footnotesize$\pm$0.37 &
  93.59\footnotesize$\pm$0.60 &
  93.77\footnotesize$\pm$0.52 &
  92.32\footnotesize$\pm$0.19 &
  94.01\footnotesize$\pm$0.33 &
   92.26\footnotesize$\pm$0.19&
   94.50\footnotesize$\pm$0.29&
   94.90\footnotesize$\pm$0.30&
   \bf95.67\footnotesize$\pm$0.40&
   \underline{95.38\footnotesize$\pm$0.44}
   \\
 &
   &
  80\% &
  89.45\footnotesize$\pm$0.36 &
  91.77\footnotesize$\pm$0.27 &
  92.58\footnotesize$\pm$0.39 &
  93.65\footnotesize$\pm$0.52 &
  92.73\footnotesize$\pm$0.66 &
  94.36\footnotesize$\pm$0.43 &
  94.46\footnotesize$\pm$0.60 &
  93.46\footnotesize$\pm$0.55 &
  94.25\footnotesize$\pm$0.57 &
   93.13\footnotesize$\pm$0.24&
  94.67\footnotesize$\pm$0.12&
  94.77\footnotesize$\pm$0.19&
  \underline{95.29\footnotesize$\pm$0.15}&
  \bf95.70\footnotesize$\pm$0.29
   \\ \cline{2-17}  
 &
  \multirow{4}{*}{Micro-F1} &
  20\% &
  88.21\footnotesize$\pm$0.26 &
  91.29\footnotesize$\pm$0.31 &
  92.06\footnotesize$\pm$0.33 &
  93.32\footnotesize$\pm$0.44 &
  92.35\footnotesize$\pm$0.51 &
  93.60\footnotesize$\pm$0.59 &
  93.15\footnotesize$\pm$0.48 &
  91.46\footnotesize$\pm$0.77 &
  94.17\footnotesize$\pm$0.47 &
   92.01\footnotesize$\pm$0.53&
   94.53\footnotesize$\pm$0.17&
   94.66\footnotesize$\pm$0.07&
  \underline{95.79\footnotesize$\pm$0.16} &
  \bf95.82\footnotesize$\pm$0.31
   \\
 &
   &
  40\% &
  88.68\footnotesize$\pm$0.52 &
  91.60\footnotesize$\pm$0.50 &
  92.31\footnotesize$\pm$0.40 &
  93.74\footnotesize$\pm$0.56 &
  92.87\footnotesize$\pm$0.39 &
  93.75\footnotesize$\pm$0.44 &
  93.80\footnotesize$\pm$0.56 &
  92.05\footnotesize$\pm$0.48 &
  93.87\footnotesize$\pm$0.42 &
   92.73\footnotesize$\pm$0.51&
   94.60\footnotesize$\pm$0.22&
   94.72\footnotesize$\pm$0.10&
  \underline{95.90\footnotesize$\pm$0.11}&
  \bf96.07\footnotesize$\pm$0.50
   \\
 &
   &
  60\% &
  90.01\footnotesize$\pm$0.48 &
  92.09\footnotesize$\pm$0.41 &
  93.16\footnotesize$\pm$0.36 &
  93.98\footnotesize$\pm$0.69 &
  93.42\footnotesize$\pm$0.12 &
  94.20\footnotesize$\pm$0.51 &
  94.22\footnotesize$\pm$0.51 &
  92.72\footnotesize$\pm$0.24 &
  94.71\footnotesize$\pm$0.56 &
   93.50\footnotesize$\pm$0.26&
   94.92\footnotesize$\pm$0.35&
   95.15\footnotesize$\pm$0.36&
   \bf95.98\footnotesize$\pm$0.36&
   \underline{95.70\footnotesize$\pm$0.42}
   \\
 &
   &
  80\% &
  90.14\footnotesize$\pm$0.39 &
  92.39\footnotesize$\pm$0.41 &
  93.21\footnotesize$\pm$0.35 &
  94.30\footnotesize$\pm$0.71 &
  93.54\footnotesize$\pm$0.60 &
  94.09\footnotesize$\pm$0.52 &
  94.23\footnotesize$\pm$0.54 &
  92.57\footnotesize$\pm$0.72 &
  94.68\footnotesize$\pm$0.55 &
   93.31\footnotesize$\pm$0.12&
   95.36\footnotesize$\pm$0.23&
   95.34\footnotesize$\pm$0.17&
  \underline{95.85\footnotesize$\pm$0.16}&
  \bf96.05\footnotesize$\pm$0.34
   \\ \hline
\multirow{8}{*}{ACM} &
  \multirow{4}{*}{Macro-F1} &
  20\% &
  83.08\footnotesize$\pm$0.37 &
  86.14\footnotesize$\pm$0.49 &
  87.29\footnotesize$\pm$1.06 &
  88.11\footnotesize$\pm$0.94 &
  87.88\footnotesize$\pm$0.42 &
  88.43\footnotesize$\pm$0.51 &
  91.10\footnotesize$\pm$0.39 &
  89.12\footnotesize$\pm$0.46 &
  92.25\footnotesize$\pm$0.39 &
   89.86\footnotesize$\pm$0.83&
   92.56\footnotesize$\pm$0.21&
   91.34\footnotesize$\pm$0.39&
  \underline{92.73\footnotesize$\pm$0.52} &
  \bf93.86\footnotesize$\pm$0.45
   \\
 &
   &
  40\% &
  87.34\footnotesize$\pm$0.41 &
  87.11\footnotesize$\pm$0.22 &
  89.19\footnotesize$\pm$0.72 &
  89.87\footnotesize$\pm$0.85 &
  90.54\footnotesize$\pm$0.08 &
  90.16\footnotesize$\pm$0.91&
  91.34\footnotesize$\pm$0.44 &
  89.15\footnotesize$\pm$0.49 &
  92.64\footnotesize$\pm$0.61 &
   90.52\footnotesize$\pm$0.69&
   92.88\footnotesize$\pm$0.19&
   92.92\footnotesize$\pm$0.32&
  \underline{93.95\footnotesize$\pm$0.51}&
  \bf94.85\footnotesize$\pm$0.18
  
   \\
 &
   &
  60\% &
  88.80\footnotesize$\pm$0.51 &
  88.92\footnotesize$\pm$0.36 &
  90.01\footnotesize$\pm$0.42 &
  91.36\footnotesize$\pm$0.59 &
  91.22\footnotesize$\pm$0.36 &
  90.73\footnotesize$\pm$0.39 &
  91.34\footnotesize$\pm$0.26 &
  90.57\footnotesize$\pm$0.34 &
  93.06\footnotesize$\pm$0.22 &
   91.03\footnotesize$\pm$0.76&
   94.10\footnotesize$\pm$0.37&
   94.28\footnotesize$\pm$0.35&
   \underline{94.83\footnotesize$\pm$0.16}&
  \bf 95.16\footnotesize$\pm$0.47
   \\
 &
   &
  80\% &
  88.43\footnotesize$\pm$0.29 &
  88.06\footnotesize$\pm$0.16 &
  90.03\footnotesize$\pm$0.77 &
  91.57\footnotesize$\pm$0.76 &
  91.35\footnotesize$\pm$0.45 &
  92.12\footnotesize$\pm$0.51 &
  91.14\footnotesize$\pm$0.78 &
  93.45\footnotesize$\pm$0.65 &
  93.55\footnotesize$\pm$0.44 &
   91.97\footnotesize$\pm$0.55&
  \bf94.94\footnotesize$\pm$0.62&
   93.91\footnotesize$\pm$0.31&
   94.01\footnotesize$\pm$0.30 &
   \underline{94.89\footnotesize$\pm$0.26}
   \\ \cline{2-17}  
 &
  \multirow{4}{*}{Micro-F1} &
  20\% &
  87.75\footnotesize$\pm$0.33 &
  87.83\footnotesize$\pm$0.47 &
  88.09\footnotesize$\pm$0.89 &
  89.36\footnotesize$\pm$0.81 &
  91.20\footnotesize$\pm$0.46 &
  91.37\footnotesize$\pm$0.42 &
  91.86\footnotesize$\pm$0.40 &
  89.59\footnotesize$\pm$0.37 &
  91.91\footnotesize$\pm$0.33 &
   90.21\footnotesize$\pm$0.61&
   92.38\footnotesize$\pm$0.18&
   92.36\footnotesize$\pm$0.37&
 \underline{92.96\footnotesize$\pm$0.54}&
 \bf93.83\footnotesize$\pm$0.49
   \\
 &
   &
  40\% &
  87.86\footnotesize$\pm$0.42 &
  87.39\footnotesize$\pm$0.41 &
  90.06\footnotesize$\pm$0.73 &
  89.91\footnotesize$\pm$0.92 &
  91.78\footnotesize$\pm$0.28 &
  92.60\footnotesize$\pm$0.48 &
  91.89\footnotesize$\pm$0.46 &
  90.70\footnotesize$\pm$0.43 &
  92.86\footnotesize$\pm$0.84 &
   90.63\footnotesize$\pm$0.52&
   93.37\footnotesize$\pm$0.26&
   93.46\footnotesize$\pm$0.50&
   \underline{93.91\footnotesize$\pm$0.48}&
   \bf94.82\footnotesize$\pm$0.22
   \\
 &
   &
  60\% &
  88.40\footnotesize$\pm$0.56 &
  87.78\footnotesize$\pm$0.33 &
  90.51\footnotesize$\pm$0.63 &
  90.66\footnotesize$\pm$0.60 &
  92.39\footnotesize$\pm$0.42 &
  92.21\footnotesize$\pm$0.18 &
  92.07\footnotesize$\pm$0.48 &
  91.18\footnotesize$\pm$0.15 &
  93.33\footnotesize$\pm$0.21 &
   91.20\footnotesize$\pm$0.48&
   94.46\footnotesize$\pm$0.35&
   94.34\footnotesize$\pm$0.39&
   \underline{94.88\footnotesize$\pm$0.16}&
   \bf95.47\footnotesize$\pm$0.56
   \\
 &
   &
  80\% &
  88.56\footnotesize$\pm$0.33 &
  87.87\footnotesize$\pm$0.51 &
  91.10\footnotesize$\pm$0.44 &
  92.41\footnotesize$\pm$0.68 &
  92.03\footnotesize$\pm$0.16 &
  92.14\footnotesize$\pm$0.48 &
  92.21\footnotesize$\pm$0.66 &
  91.77\footnotesize$\pm$0.57 &
  93.53\footnotesize$\pm$0.42 &
   92.06\footnotesize$\pm$0.66&
  \underline{94.56\footnotesize$\pm$0.11}&
   93.72\footnotesize$\pm$0.32&
   94.05\footnotesize$\pm$0.31&
   \bf94.96\footnotesize$\pm$0.15   
   \\ \hline
\end{tabular}%
}
\end{table*}

\begin{table*}[]
\centering
\caption{Experimental results (\%) for the node clustering task. The best and second-best performers are bolded and underlined, respectively.}
\label{tab:node_clustering}
\resizebox{\textwidth}{!}{%
\begin{tabular}{cccccccccccccccc}
\hline
\multirow{2}{*}{Dataset} &
  \multirow{2}{*}{Metric} &
  \multicolumn{2}{c}{\lowercase\expandafter{\romannumeral1}} & \multicolumn{2}{c}{\lowercase\expandafter{\romannumeral2}} & \multicolumn{5}{c}{\lowercase\expandafter{\romannumeral3}} & \multicolumn{5}{c}{\lowercase\expandafter{\romannumeral4}}\\ \cmidrule(lr){3-4} \cmidrule(lr){5-6} \cmidrule(lr){7-11} \cmidrule(lr){12-16}
  &
   &
  GCN &
  GAT &
  HGCN &
  HGCL &
  HAN &
  MAGNN &
  GTN &
  HGT &
  Simple-HGN &
  McH-HGCN &
  SHAN &
  HHGAT &
  MSGAT &
  \algname{} \\ \hline
\multirow{2}{*}{IMDB} &
  NMI &
  7.84\footnotesize$\pm$0.24 &
  8.06\footnotesize$\pm$0.18 &
  10.29\footnotesize$\pm$0.76 &
  12.89\footnotesize$\pm$0.93 &
  11.21\footnotesize$\pm$1.09 &
  15.66\footnotesize$\pm$0.73 &
  15.01\footnotesize$\pm$0.11 &
  14.55\footnotesize$\pm$0.32 &
  17.58\footnotesize$\pm$0.82 &
  14.32\footnotesize$\pm$0.46 &
  20.60\footnotesize$\pm$0.92 &
  20.75\footnotesize$\pm$0.36&
  \underline{24.06\footnotesize$\pm$0.51}&
   \bf26.18\footnotesize$\pm$0.67\\
 &
  ARI &
  8.12\footnotesize$\pm$0.40 &
  8.86\footnotesize$\pm$0.09 &
  11.10\footnotesize$\pm$0.88 &
  14.78\footnotesize$\pm$0.73 &
  11.49\footnotesize$\pm$0.11 &
  16.72\footnotesize$\pm$0.21 &
  15.96\footnotesize$\pm$0.63 &
  16.59\footnotesize$\pm$0.36 &
  19.51\footnotesize$\pm$1.06 &
  16.91\footnotesize$\pm$0.37 &
  22.56\footnotesize$\pm$0.22 &
  22.80\footnotesize$\pm$0.68 &
  \underline{26.33\footnotesize$\pm$0.46}&
   \bf30.87\footnotesize$\pm$0.63\\ \hline
\multirow{2}{*}{DBLP} &
  NMI &
  75.37\footnotesize$\pm$0.25 &
  75.46\footnotesize$\pm$0.44 &
  76.48\footnotesize$\pm$0.87 &
  78.06\footnotesize$\pm$0.43 &
  77.03\footnotesize$\pm$0.16 &
  80.11\footnotesize$\pm$0.30 &
  81.39\footnotesize$\pm$0.73 &
  79.02\footnotesize$\pm$0.39 &
  82.38\footnotesize$\pm$0.07 &
  78.90\footnotesize$\pm$0.31 &
  82.39\footnotesize$\pm$0.42 &
  83.14\footnotesize$\pm$0.19 &
  \underline{84.38\footnotesize$\pm$0.59}&
   \bf84.52\footnotesize$\pm$0.34\\
 &
  ARI &
  77.14\footnotesize$\pm$0.21 &
  77.99\footnotesize$\pm$0.72 &
  79.36\footnotesize$\pm$0.95 &
  80.49\footnotesize$\pm$0.61 &
  82.53\footnotesize$\pm$0.42 &
  85.61\footnotesize$\pm$0.38 &
  84.12\footnotesize$\pm$0.83 &
  80.28\footnotesize$\pm$0.20 &
  85.71\footnotesize$\pm$0.33 &
  81.22\footnotesize$\pm$0.56 &
  86.13\footnotesize$\pm$0.33 &
  85.91\footnotesize$\pm$0.49 &
  \underline{88.27\footnotesize$\pm$0.63}&
   \bf88.89\footnotesize$\pm$0.28\\ \hline
\multirow{2}{*}{ACM} &
  NMI &
  51.73\footnotesize$\pm$0.21 &
  58.06\footnotesize$\pm$0.46 &
  60.19\footnotesize$\pm$0.69 &
  62.84\footnotesize$\pm$0.74 &
  61.24\footnotesize$\pm$0.12 &
  64.73\footnotesize$\pm$0.47 &
  65.06\footnotesize$\pm$0.35 &
  67.88\footnotesize$\pm$0.20 &
  69.91\footnotesize$\pm$0.68 &
  66.76\footnotesize$\pm$0.38 &
  72.90\footnotesize$\pm$0.93 &
  72.49\footnotesize$\pm$0.44 &
  \underline{73.33\footnotesize$\pm$0.79}&
   \bf76.62\footnotesize$\pm$0.81\\
 &
  ARI &
  53.42\footnotesize$\pm$0.48 &
  59.61\footnotesize$\pm$0.42 &
  62.06\footnotesize$\pm$0.70 &
  64.50\footnotesize$\pm$0.72 &
  64.11\footnotesize$\pm$0.26 &
  66.84\footnotesize$\pm$0.25 &
  65.80\footnotesize$\pm$0.49 &
  72.56\footnotesize$\pm$0.13 &
  72.07\footnotesize$\pm$0.51 &
  71.84\footnotesize$\pm$0.48 &
  77.73\footnotesize$\pm$0.44 &
  77.92\footnotesize$\pm$0.80 &
  \underline{78.28\footnotesize$\pm$ 1.07}&
  \bf81.73\footnotesize$\pm$0.95 \\ \hline
\end{tabular}%
}
\end{table*}

\begin{table*}[]
\centering
\caption{Experimental results (\%) for the link prediction task. The best and second-best performers are bolded and underlined, respectively.}
\label{tab:link_prediction}
\resizebox{0.8\textwidth}{!}{%
\begin{tabular}{ccccccccccccc}
\hline
\multirow{2}{*}{Dataset} &
  \multirow{2}{*}{Metric} &
  \multicolumn{2}{c}{\lowercase\expandafter{\romannumeral1}} & \multicolumn{2}{c}{\lowercase\expandafter{\romannumeral2}} & \multicolumn{4}{c}{\lowercase\expandafter{\romannumeral3}} & \multicolumn{3}{c}{\lowercase\expandafter{\romannumeral4}}\\ \cmidrule(lr){3-4} \cmidrule(lr){5-6} \cmidrule(lr){7-10} \cmidrule(lr){11-13}
  &
   &
  GCN &
  GAT &
  HGCN &
  HGCL &
  HAN &
  MAGNN &
  HGT &
  Simple-HGN &
  HHGAT &
  MSGAT &
  \algname{} \\ \hline
\multirow{2}{*}{LastFM} &
  ROC-AUC &
  43.68\footnotesize$\pm$0.30 &
  44.52\footnotesize$\pm$0.22 &
  46.71\footnotesize$\pm$0.78 &
  46.99\footnotesize$\pm$0.97 &
  48.32\footnotesize$\pm$0.28 &
  49.37\footnotesize$\pm$0.59 &
  47.78\footnotesize$\pm$0.23 &
  53.85\footnotesize$\pm$0.47 &
  54.37\footnotesize$\pm$0.51&
  \underline{55.77\footnotesize$\pm$0.62} &
  \bf56.39\footnotesize$\pm$0.84\cr
 &
  F1-Score &
  56.15\footnotesize$\pm$0.16 &
  56.84\footnotesize$\pm$0.07 &
  57.23\footnotesize$\pm$0.66 &
  58.03\footnotesize$\pm$0.83 &
  57.11\footnotesize$\pm$0.49 &
  58.37\footnotesize$\pm$0.32 &
  61.16\footnotesize$\pm$0.57 &
  63.02\footnotesize$\pm$0.35 &
  62.85\footnotesize$\pm$0.48 &
  \underline{63.39\footnotesize$\pm$0.76}&
  \bf63.48\footnotesize$\pm$0.77\cr
  \hline
\end{tabular}%
}
\end{table*}
\subsection{Implementation Details}
For the baselines including \algname{}, we randomly initialize model parameters and use the AdamW~\cite{LoshchilovH19} optimizer with a learning rate of 0.0001 and a weight decay of 0.001. We set the dropout rate to 0.5, and the dimension of final node embedding to 64. For multi-head attention models, we set the number of attention heads to 8. For \algname{}, we set the dimension of metapath embedding to 128. The baseline models are trained for 100 epochs, and we use early stopping with patience of 10 epochs. For the metapath-based heterogeneous GNNs, the metapath settings follow the specifications outlined in their papers. For \algname{}, the maximum length of metapath $l$ is set to 4, 5, 4 and 3 for IMDB, DBLP, ACM and LastFM, respectively. All downstream tasks were conducted ten times, and we report the average values of evaluation metrics along with their corresponding standard deviations.

\subsection{Evaluation Metrics}
For node classification, we use Macro-F1 and Micro-F1 scores to evaluate classification accuracy. For node clustering, Normalized Mutual Information (NMI) and Adjusted Rand Index (ARI) are used to evaluate clustering performance. For link prediction, we use the area under the ROC curve (ROC-AUC) and F1-score to measure prediction accuracy.

\subsection{Node Classification and Clustering (RQ1-1)}
Node classification was performed by applying support vector machines on embedding vectors of labeled nodes. The ratio of training data was varied within the range of 20\% to 80\%. For node clustering, the $k$-means clustering algorithm was applied to embedding vectors of labeled nodes. Node classification and clustering performed ten times, and we report average values and standard deviations of metrics for node classification and clustering.

As shown in Table~\ref{tab:node_classification} and~\ref{tab:node_clustering}, \algname{} outperforms other baselines in most cases. 

Additionally, we would like to provide three empirical findings. First, the results of \algname{} and MSGAT demonstrate the effectiveness of the proposed metapath-based hyperbolic contrastive learning method. Specifically, although MSGAT achieves significant improvements by leveraging multiple hyperbolic spaces to capture various complex structures within a heterogeneous graph, it restricts its ability to effectively capture semantic differences among diverse metapaths. In contrast, by employing a metapath-based hyperbolic contrastive learning method, \algname{} enhances the discriminability of embeddings derived from distinct metapaths. By learning compact and well-separated representations for each metapath, \algname{} more effectively captures the inherent semantic diversity within heterogeneous graphs. 

Second, in comparison with HGCN and HGCL, both models use the hyperbolic space for graph representation learning. However, while HGCN captures hierarchical structures more effectively than Euclidean homogeneous GNNs (i.e., GCN, GAT), HGCL achieves better performance by adopting a hyperbolic contrastive learning method, which more effectively leverages the geometric properties of the hyperbolic space. Moreover, a comparison of \algname{} and HGCL demonstrates that \algname{} effectively captures the heterogeneous information by leveraging metapaths within heterogeneous graphs. In contrast, HGCL, originally proposed for homogeneous graph representation learning, is not effective in capturing such information. Similarly, when comparing hyperbolic homogeneous GNNs and Euclidean heterogeneous GNNs, we observe that although hyperbolic homogeneous GNNs are more effective in capturing complex structures, they are unable to learn heterogeneous semantic information. In contrast, Euclidean heterogeneous GNNs can effectively model such information by considering multiple types of nodes and links.

Finally, when comparing Euclidean heterogeneous GNNs and hyperbolic heterogeneous GNNs, we can conclude that hyperbolic heterogeneous GNNs are more effective as they can simultaneously capture the complex structures and heterogeneity of heterogeneous graphs.

\begin{table*}[]
\centering
\caption{Results of the ablation study.}
\label{tab:ablation}
\resizebox{\textwidth}{!}{%
\begin{tabular}{ccccccccccccc}
\hline
Dataset & \multicolumn{4}{c}{IMDB} & \multicolumn{4}{c}{DBLP} & \multicolumn{4}{c}{ACM} \\ \cmidrule(lr){2-5} \cmidrule(lr){6-9} \cmidrule(lr){10-13}
Metric & Macro-F1 & Micro-F1 & NMI & ARI & Macro-F1 & Micro-F1 & NMI & ARI & Macro-F1 & Micro-F1 & NMI & ARI \\ \hline
\algname{} & 
\bf72.29\footnotesize$\pm$0.59 & 
\bf74.63\footnotesize$\pm$0.80 &
\bf26.18\footnotesize$\pm$0.67 &
\bf30.87\footnotesize$\pm$0.63 &
\bf95.38\footnotesize$\pm$0.44 & 
\bf95.70\footnotesize$\pm$0.42 &
\bf84.52\footnotesize$\pm$0.34 & 
\bf88.89\footnotesize$\pm$0.28 &
\bf95.16\footnotesize$\pm$0.47 & 
\bf95.47\footnotesize$\pm$0.56 &
\bf76.62\footnotesize$\pm$0.81 &
\bf81.73\footnotesize$\pm$0.95    
\\
\algname{}$_{\text{w/o cont}}$ &
70.92\footnotesize$\pm$0.79 & 
73.18\footnotesize$\pm$0.86 & 
23.90\footnotesize$\pm$0.66 & 
27.70\footnotesize$\pm$0.53 & 
94.90\footnotesize$\pm$0.71 & 
95.15\footnotesize$\pm$0.74 & 
83.69\footnotesize$\pm$0.62 & 
87.84\footnotesize$\pm$0.60 & 
94.60\footnotesize$\pm$0.10 & 
94.53\footnotesize$\pm$0.08 & 
73.85\footnotesize$\pm$0.41 & 
78.01\footnotesize$\pm$0.44
\\
\algname{}$_{\text{Single+Cont}}$ &
67.88\footnotesize$\pm$0.48 & 
69.49\footnotesize$\pm$0.36 &
23.73\footnotesize$\pm$0.51 & 
28.63\footnotesize$\pm$0.49 & 
94.42\footnotesize$\pm$0.25 & 
94.85\footnotesize$\pm$0.40 & 
80.26\footnotesize$\pm$0.76 & 
86.38\footnotesize$\pm$0.52 & 
93.23\footnotesize$\pm$0.13 & 
93.20\footnotesize$\pm$0.17 & 
74.16\footnotesize$\pm$0.48 & 
78.85\footnotesize$\pm$0.44
\\ 
\algname{}$_{\text{Single}}$ &
66.19\footnotesize$\pm$0.76 &
68.35\footnotesize$\pm$0.53 &
21.33\footnotesize$\pm$0.45 & 
25.40\footnotesize$\pm$0.74 & 
93.74\footnotesize$\pm$0.97 & 
93.86\footnotesize$\pm$0.89 & 
80.12\footnotesize$\pm$0.64 & 
86.05\footnotesize$\pm$0.71 & 
92.40\footnotesize$\pm$0.46 & 
92.46\footnotesize$\pm$0.58 & 
73.53\footnotesize$\pm$0.96 & 
78.10\footnotesize$\pm$0.82
\\ 
\algname{}$_{\text{Euclid+Cont}}$ &
66.59\footnotesize$\pm$0.46 & 
68.37\footnotesize$\pm$0.69 & 
21.79\footnotesize$\pm$0.13 & 
26.17\footnotesize$\pm$0.16 & 
93.60\footnotesize$\pm$0.52 & 
93.73\footnotesize$\pm$0.38 & 
79.75\footnotesize$\pm$0.50 & 
83.98\footnotesize$\pm$0.44 & 
91.39\footnotesize$\pm$0.81 & 
91.43\footnotesize$\pm$0.77 & 
71.04\footnotesize$\pm$0.49 & 
74.82\footnotesize$\pm$0.84 
\\
\algname{}$_{\text{Euclid}}$ & 
64.56\footnotesize$\pm$0.84 & 
67.32\footnotesize$\pm$0.71 & 
16.93\footnotesize$\pm$0.09 & 
16.16\footnotesize$\pm$0.12 & 
93.18\footnotesize$\pm$0.49 & 
93.53\footnotesize$\pm$0.33 & 
78.94\footnotesize$\pm$0.41 & 
83.68\footnotesize$\pm$0.48 & 
90.32\footnotesize$\pm$0.23 & 
90.51\footnotesize$\pm$0.22 & 
70.16\footnotesize$\pm$0.56 & 
73.31\footnotesize$\pm$0.64
\\\hline
\end{tabular}%
}
\end{table*}

\begin{table}[]
\centering
%\footnotesize
\caption{Variant models of \algname{}.}
\resizebox{\columnwidth}{!}{%
\label{tab:variants}
\begin{tabular}{c|c|c}
\hline
Models & Embedding Space & Contrastive Learning \\\hline
\algname{}$_{\text{w/o Cont}}$ & Multi hyperbolic space & \xmark \\
\algname{}$_{\text{Single+Cont}}$ & Single hyperbolic space & \cmark \\
\algname{}$_{\text{Single}}$ & Single hyperbolic space & \xmark \\
\algname{}$_{\text{Euclid+Cont}}$ & Euclidean space & \cmark \\
\algname{}$_{\text{Euclid}}$  & Euclidean space & \xmark \\\hline
\end{tabular}%
}
\end{table}

\subsection{Link Prediction (RQ1-2)}
We also conducted a link prediction task on the LastFM dataset. To predict the probabilities of relations between user-type nodes and artist-type nodes, we applied a dot product operation to the embeddings of the two types of nodes. We considered all connected user artist pairs as positive samples, while unconnected user-artist pairs were considered as negative samples. For model training, we use the same number of positive and negative samples.

As shown in Table~\ref{tab:link_prediction}, \algname{} outperforms the other baselines. In the link prediction task, when comparing \algname{} with MSGAT and HHGAT, predicting the connection between two nodes of different types involves distinct metapaths depending on each node type, and the distributions of metapath instances also differ accordingly. However, HHGAT, which uses only a single hyperbolic space, fails to capture these variations. In contrast, MSGAT and \algname{}, by employing multiple hyperbolic spaces, can flexibly adapt to various metapath instance distributions. Furthermore, \algname{} shows slightly better performance than MSGAT, as it more effectively captures semantic distinctions among metapaths starting from different node types.

\subsection{Ablation Study (RQ2)}
We compose five variants of \algname{} based on two criteria: the type of embedding spaces and the use of the metapath-based contrastive learning method, to validate the effectiveness of each component of \algname{}.
Table~\ref{tab:variants} summarizes category of variant models of \algname{} used for the ablation study. Note that the training percentage for node classification is set to 60\%. We report the results of the ablation study in Table~\ref{tab:ablation}. 

First, when comparing the Euclidean space with a single hyperbolic space, the results demonstrate that the hyperbolic space is more effective in capturing the complex structures inherent in heterogeneous graphs. Furthermore, the results indicate that the multi-hyperbolic space more effectively captures the diverse semantic and structural properties derived from distinct metapaths than the single hyperbolic space case.

Second, models that use the proposed contrastive learning method consistently outperform those that do not. Notably, in some cases—such as the comparison between \algname{}$_{\text{Euclid+Cont}}$ and \algname{}$_{\text{Single}}$—although \algname{}$_{\text{Single}}$ uses a single hyperbolic space to better capture complex structures within heterogeneous graphs, \algname{}$_{\text{Euclid+Cont}}$, which uses the metapath-based contrastive learning method, achieves better performance. These results demonstrate that the proposed metapath-based contrastive learning method improves the discriminability of embeddings derived from different metapaths, leading to more effective learning of the diverse semantic information inherent in heterogeneous graphs.

Finally, \algname{}$_{\text{Single}}$ and \algname{}$_{\text{w/o cont}}$ correspond to our earlier works HHGAT~\cite{park2024hyperbolic} and MSGAT~\cite{park2024multi}, respectively. Unlike these previous models, \algname{} not only learns complex structures from metapaths, but also enhances the semantic discriminability of metapath embeddings through a metapath-based contrastive learning method. Empirically, this contrastive learning method leads to performance gains, demonstrating its effectiveness in learning well-separated metapath embeddings.

\begin{table}[t!]
\centering
\caption{Statistics of synthetic heterogeneous graphs.}
\label{tab:synthetic}
\resizebox{0.8\columnwidth}{!}{%
\begin{tabular}{ccccc} \hline
Dataset & \# Nodes & \# Links & \# Classes & \# Features \\\hline
Synthetic    & \begin{tabular}[c]{@{}c@{}}A : 6,000\\ B : 2,000\\ C : 1,000\end{tabular}   & \begin{tabular}[c]{@{}c@{}}A-B : 13,452\\ A-C : 3,954\end{tabular} & 3  & 2,000\\\hline
\end{tabular}%
}
\end{table}

\begin{figure}[t!]
\captionsetup[subfigure]{justification=centering}
\centering
    \begin{minipage}[b]{0.49\columnwidth}
        \centering
        \includegraphics[width=\linewidth]{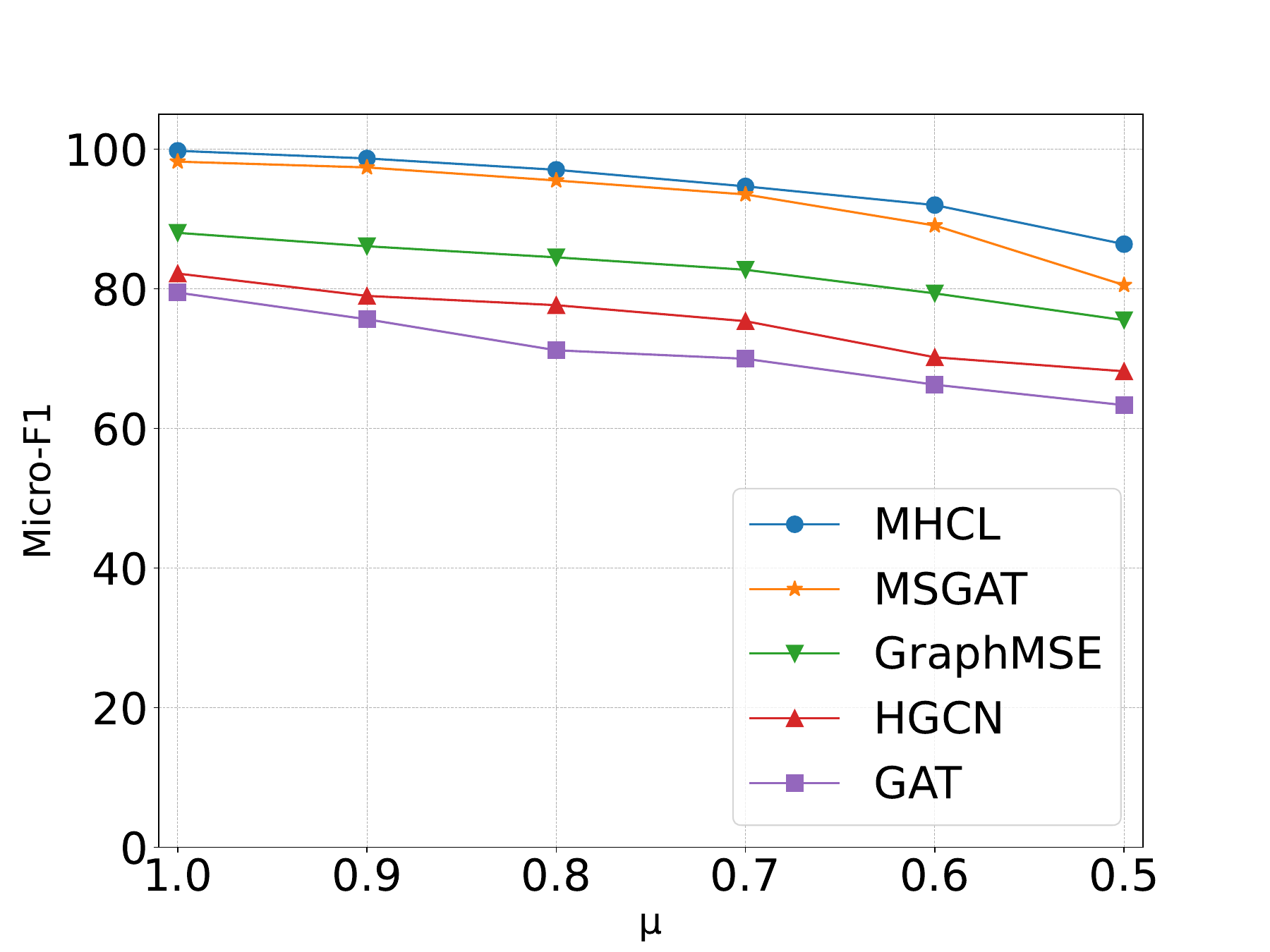}
        \subcaption{Micro-F1}
    \end{minipage}
    \begin{minipage}[b]{0.49\linewidth}
        \centering
        \includegraphics[width=\linewidth]{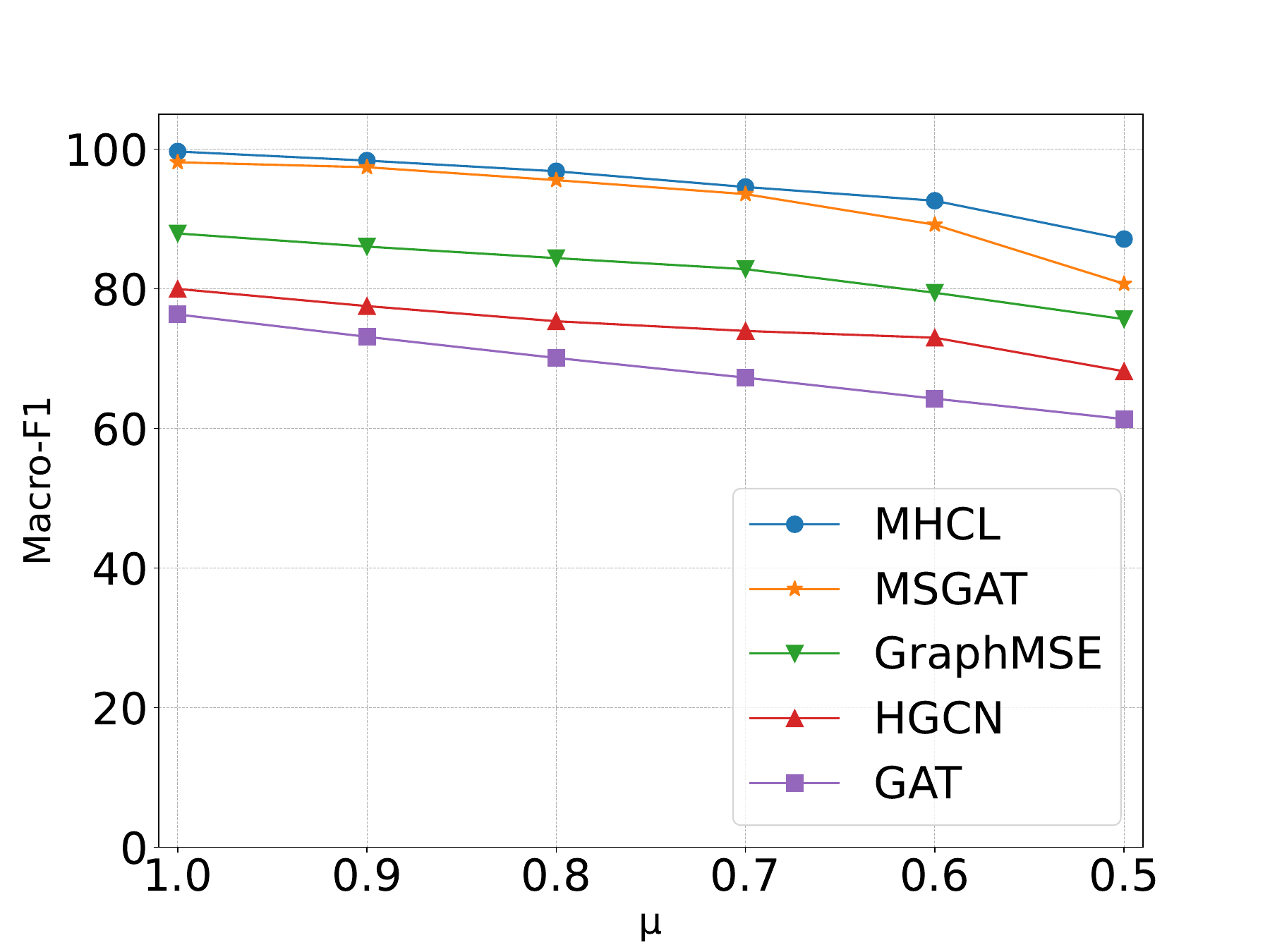}
        \subcaption{Macro-F1}
    \end{minipage}
\caption{Node classification accuracy for varying values of $\mu$.}
\label{figure:robustness}
\end{figure}

%In summary, from the results of the ablation study, we can see that using multi-hyperbolic space enables \algname{} to learn the structural distributions of complex structures from distinct metapaths, and that using metapath-based contrastive learning method further enhances their semantic discriminability. 

\begin{figure*}[t!]
\captionsetup[subfigure]{justification=centering}
\centering
    \begin{minipage}[b]{0.24\linewidth}
        \centering
        \includegraphics[width=\linewidth]{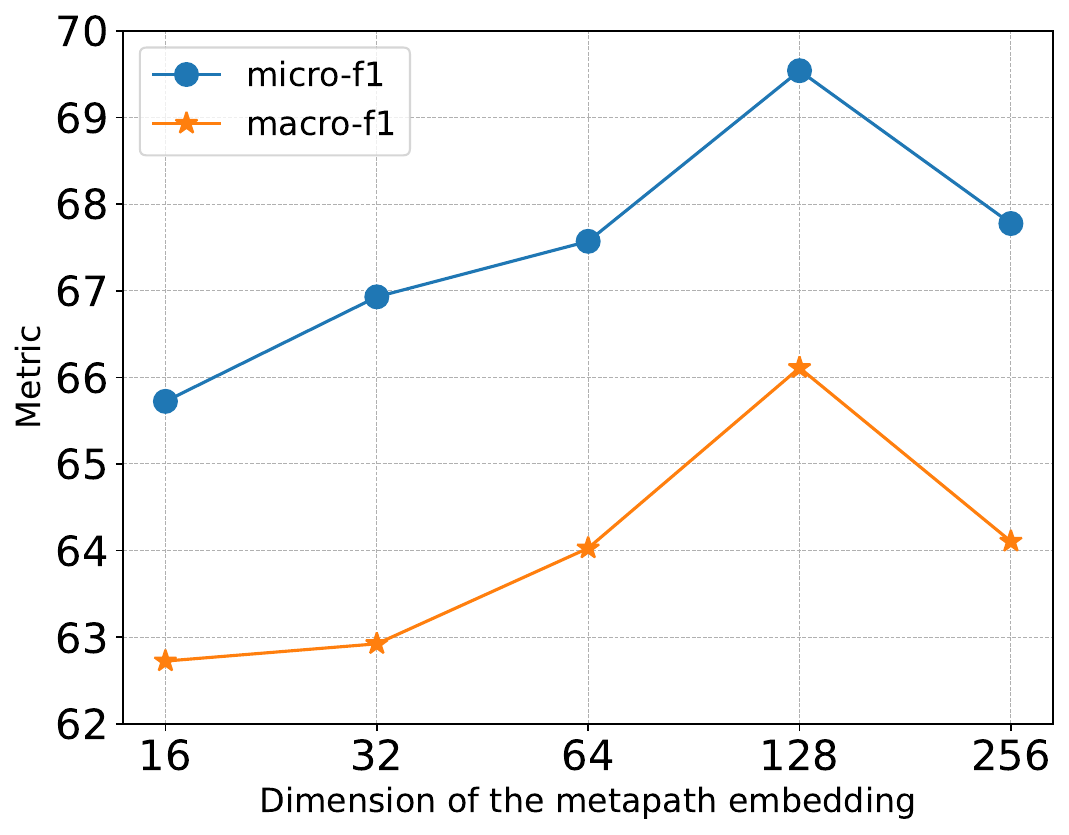}
        \subcaption{Dimension of\\ metapath embedding
        $h_v^{\phi}$}
    \end{minipage}
    \hfill
    \begin{minipage}[b]{0.24\linewidth}
        \centering
        \includegraphics[width=\linewidth]{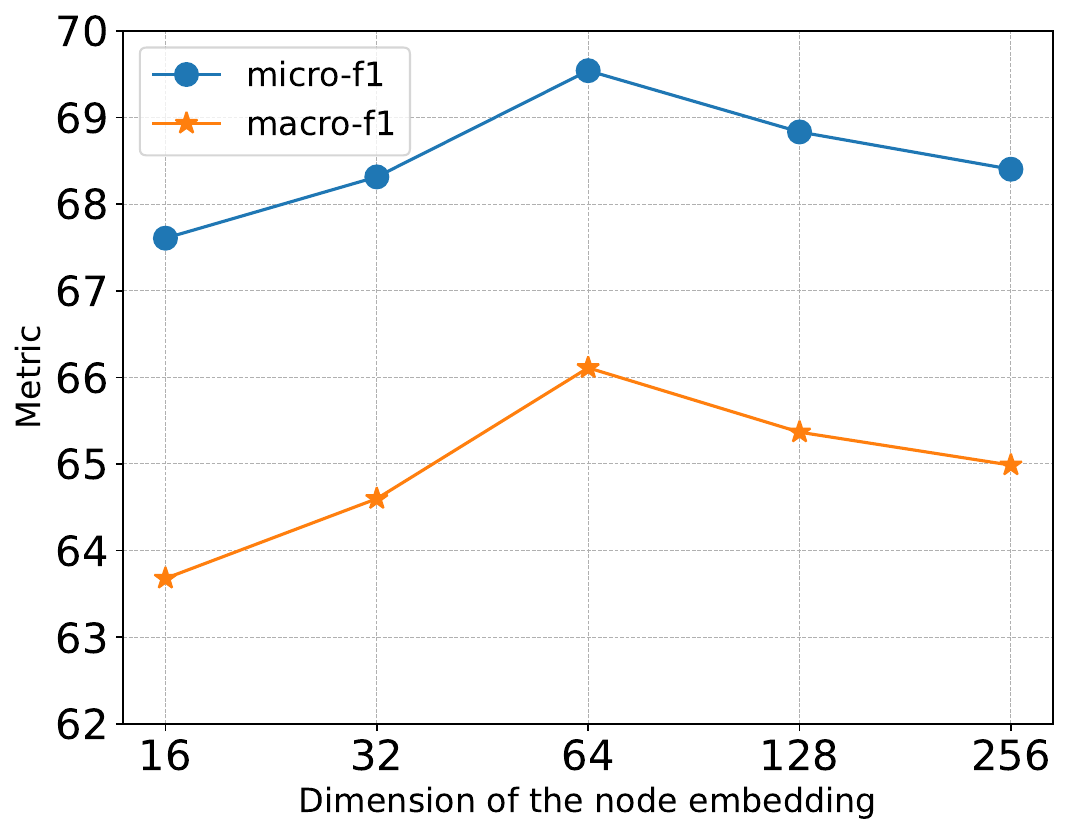}
        \subcaption{Dimension of\\ node embedding $z_v$}
    \end{minipage}
    \begin{minipage}[b]{0.24\linewidth}
        \centering
        \includegraphics[width=\linewidth]{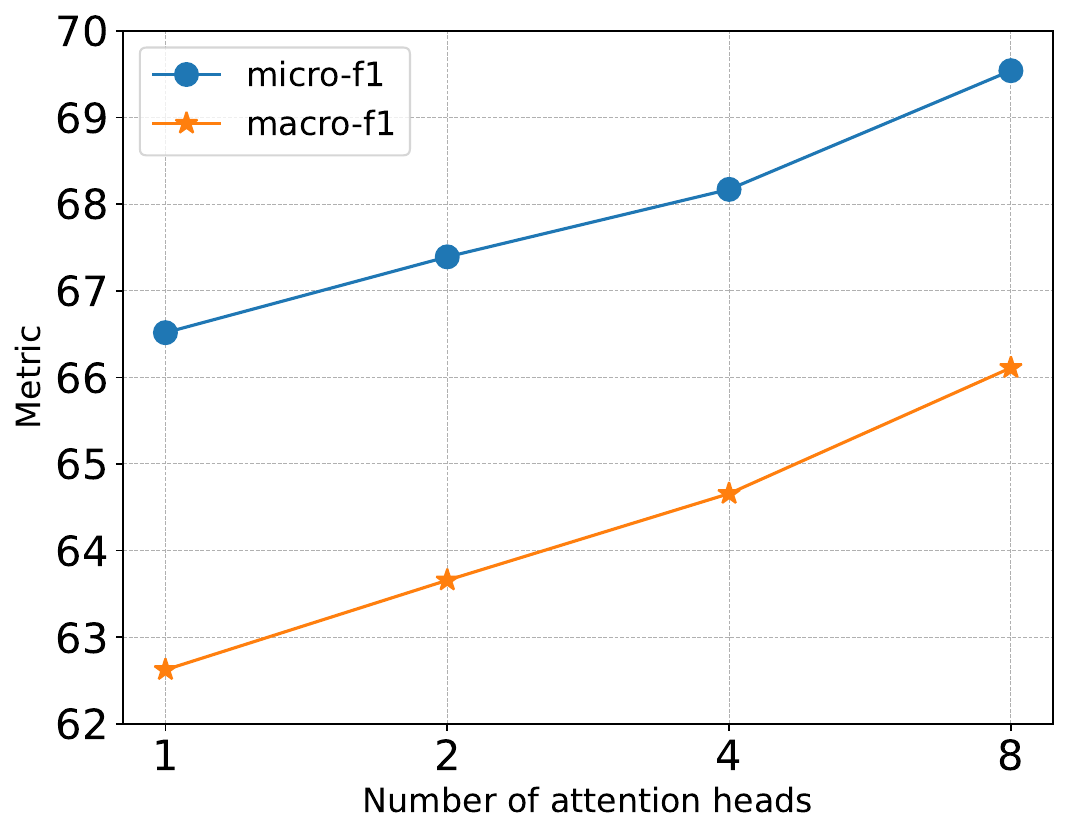}
        \subcaption{Number of attention heads\\ $K$}
    \end{minipage}
    \begin{minipage}[b]{0.24\linewidth}
        \centering
        \includegraphics[width=\linewidth]{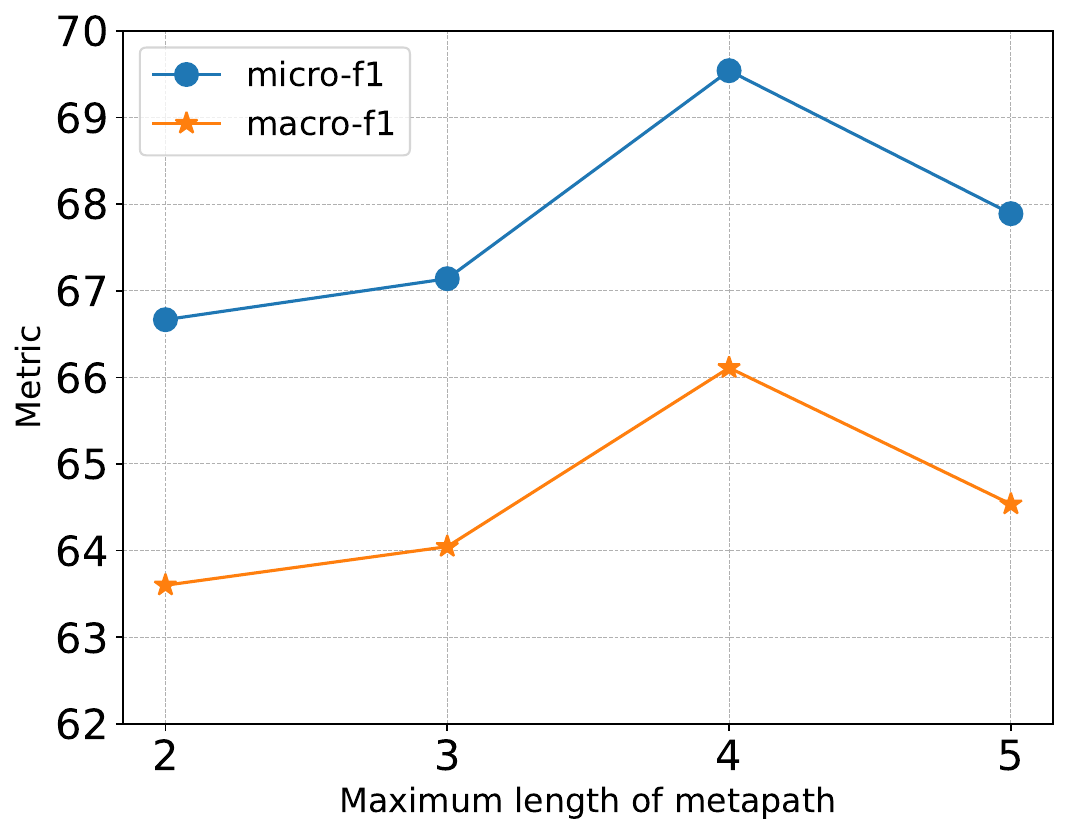}
        \subcaption{Maximum length of metapath\\ $l$}
    \end{minipage}
\caption{Parameter sensitivity of \algname{} w.r.t Dimension of the metapath embedding $h_v^{\phi}$, Dimension of the node embedding $z_v$, Number of the attention heads $K$ and Length of maximum metapath $l$.}
\label{figure:hyperparameter_study}
\end{figure*}

\subsection{Robustness Analysis (RQ3)}
To analyze the robustness of \algname{}, we generate a synthetic heterogeneous graph using the Barabasi-Albert model~\cite{barabasi1999emergence}. First, we generate a power-law distributed homogeneous graph with the Barabasi-Albert model. Then, to give heterogeneity, we randomly assign node types to all nodes within this homogeneous graph, resulting in a heterogeneous graph. Finally, the generated synthetic heterogeneous graph consists of three node types \{A, B, C\} and two link types \{AB, AC\}. Table~\ref{tab:synthetic} provides the statistics of a generated synthetic heterogeneous graph.

For node classification, we label A-type nodes into three classes. We first assume that the node features of each class follow a Gaussian distribution and generate node features according to three different Gaussian distributions $\left( \mathcal{N}(-\mu, \sigma^2),\, \mathcal{N}(0, \sigma^2),\, \mathrm{and} \; \mathcal{N}(\mu, \sigma^2) \right)$. To control the mixture between classes, we adjust the value of $\mu$ and measure the performance of \algname{} and other baselines. Note that the lower $\mu$ value indicates more severe mixing between classes.

As shown in Figure~\ref{figure:robustness}, \algname{} achieves superior performance compared to the other baselines. Although the degree of class mixing increases, the performance of \algname{} does not significantly decrease compared to the other baselines. The results demonstrate that \algname{} is more robust than the other baselines because \algname{} effectively captures not only the node features, but also the structural properties of the heterogeneous graph.

\subsection{Hyperparameter Sensitivity Analysis (RQ4)}
We investigate the impact of hyperparameters used in \algname{} and report the performance on node classification using the IMDB dataset with a training ratio of 20\%.

Figure~\ref{figure:hyperparameter_study}(a) shows that, as the dimension of the metapath embedding increases, the performance of \algname{} also increases, reaching its peak when the dimension is 128. Beyond this point, the performance of \algname{} starts to decrease. This observation demonstrates that, if the dimension is too small, the model tends to underfit, whereas, if the dimension is too large, then the model tends to overfit. Figure~\ref{figure:hyperparameter_study}(b) shows that, as the dimension of the node embedding increases, the performance of \algname{} improves, reaching its best performance when the dimension is 64. However, beyond this peak, the performance decreases steadily. Figure~\ref{figure:hyperparameter_study}(c) shows that the performance of \algname{} increases as the number of heads increases. This result demonstrates that multi-head attention can enhance the stability of the learning process. Figure~\ref{figure:hyperparameter_study}(d) shows that a maximum metapath length of four is optimal for the IMDB dataset. This finding demonstrates that a short maximum metapath length may be insufficient to capture the surrounding information of a node, while an overly long maximum metapath length may result in redundancy in structural information.

\begin{figure*}[t!]
        \centering
        \begin{minipage}[b]{0.24\linewidth}
            \centering
            \includegraphics[width=\linewidth]{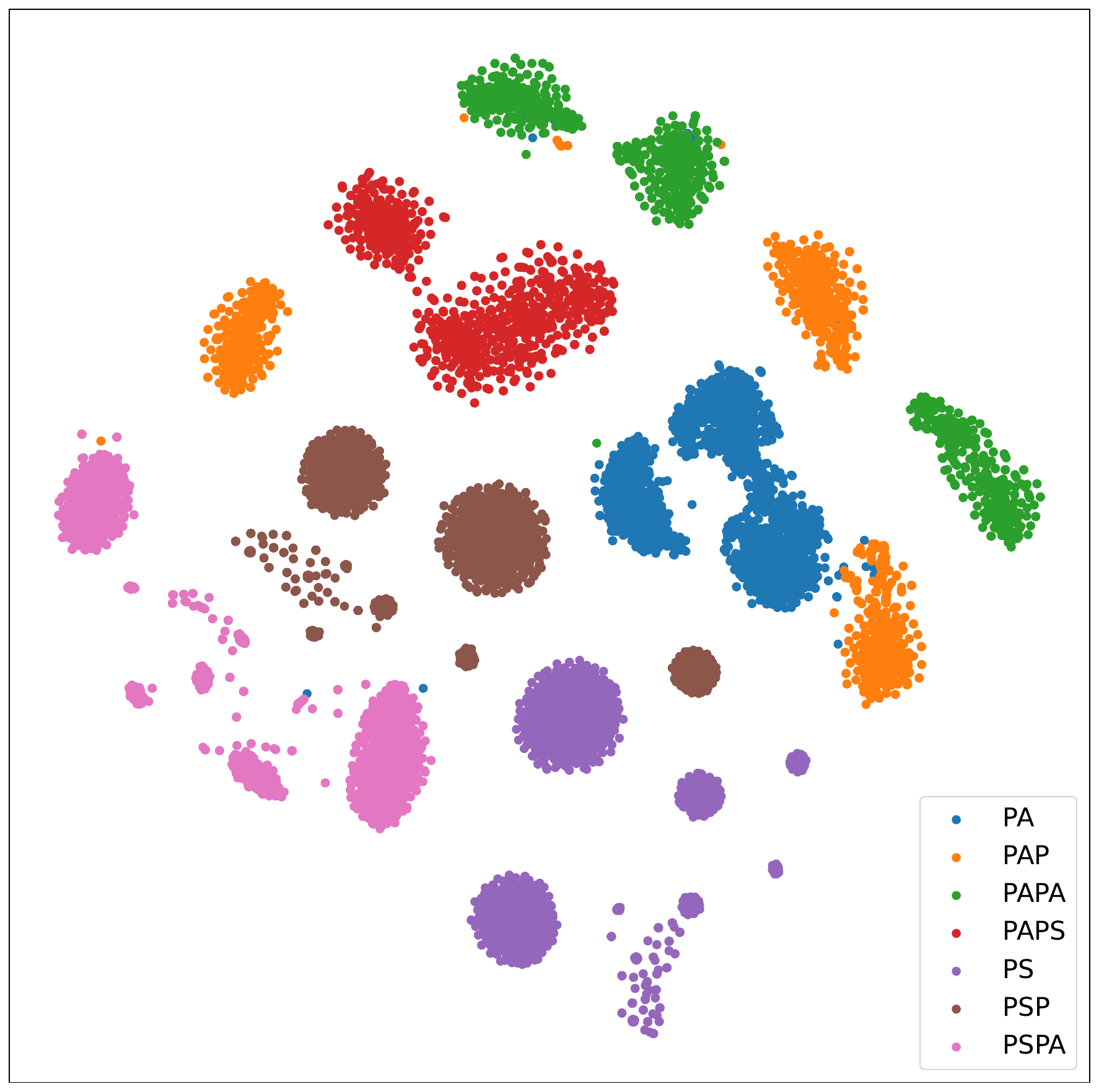}
            \subcaption{ACM w/ \algname{}}%$\mathcal{L}_{hyp}$
        \end{minipage}
        \centering
        \begin{minipage}[b]{0.24\linewidth}
            \centering
            \includegraphics[width=\linewidth]{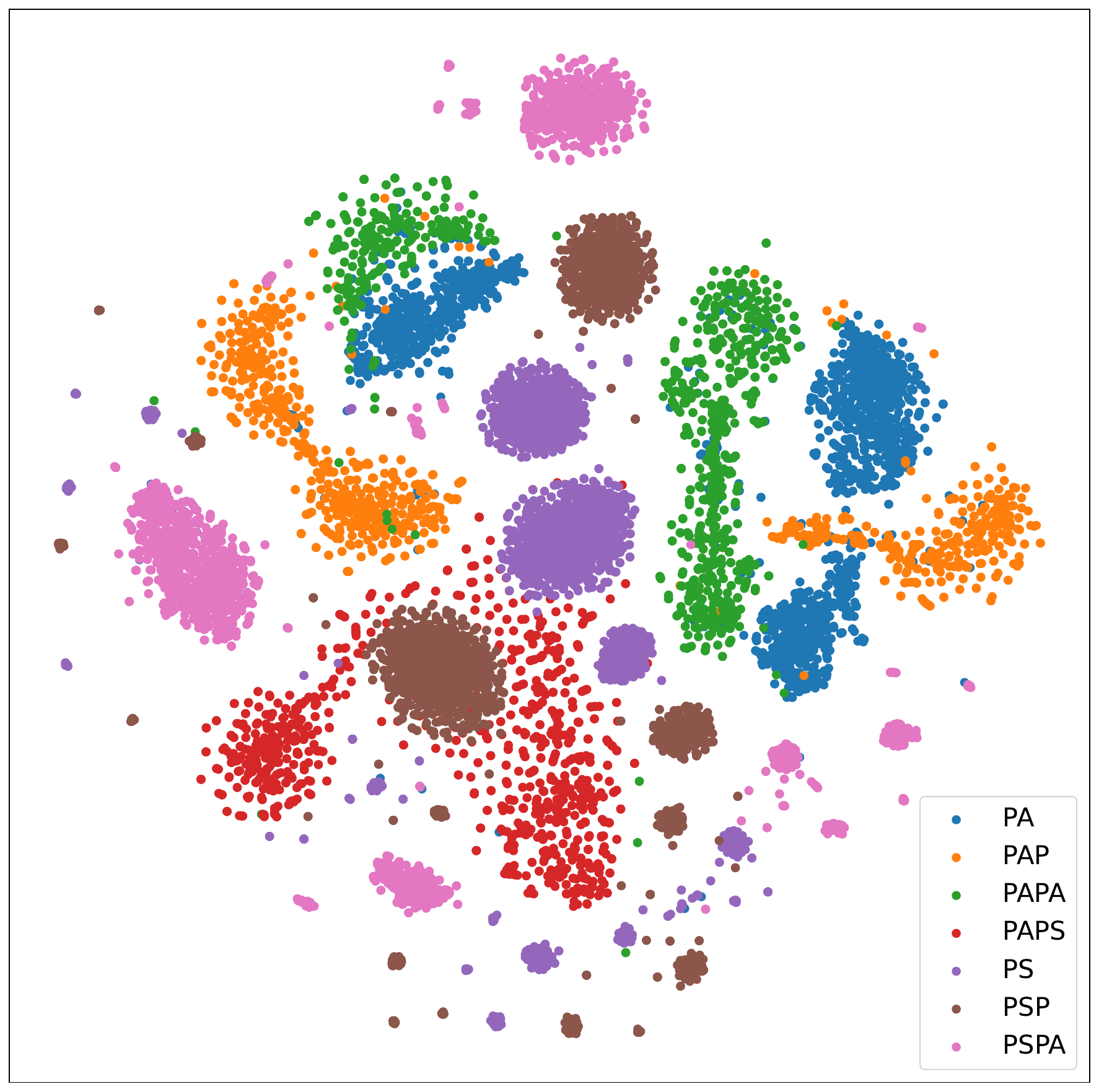}
            \subcaption{ACM w/o \algname{}}%$\mathcal{L}_{hyp}$
        \end{minipage}\centering
        %\vskip\baselineskip
        \begin{minipage}[b]{0.24\linewidth}
            \centering
            \includegraphics[width=\linewidth]{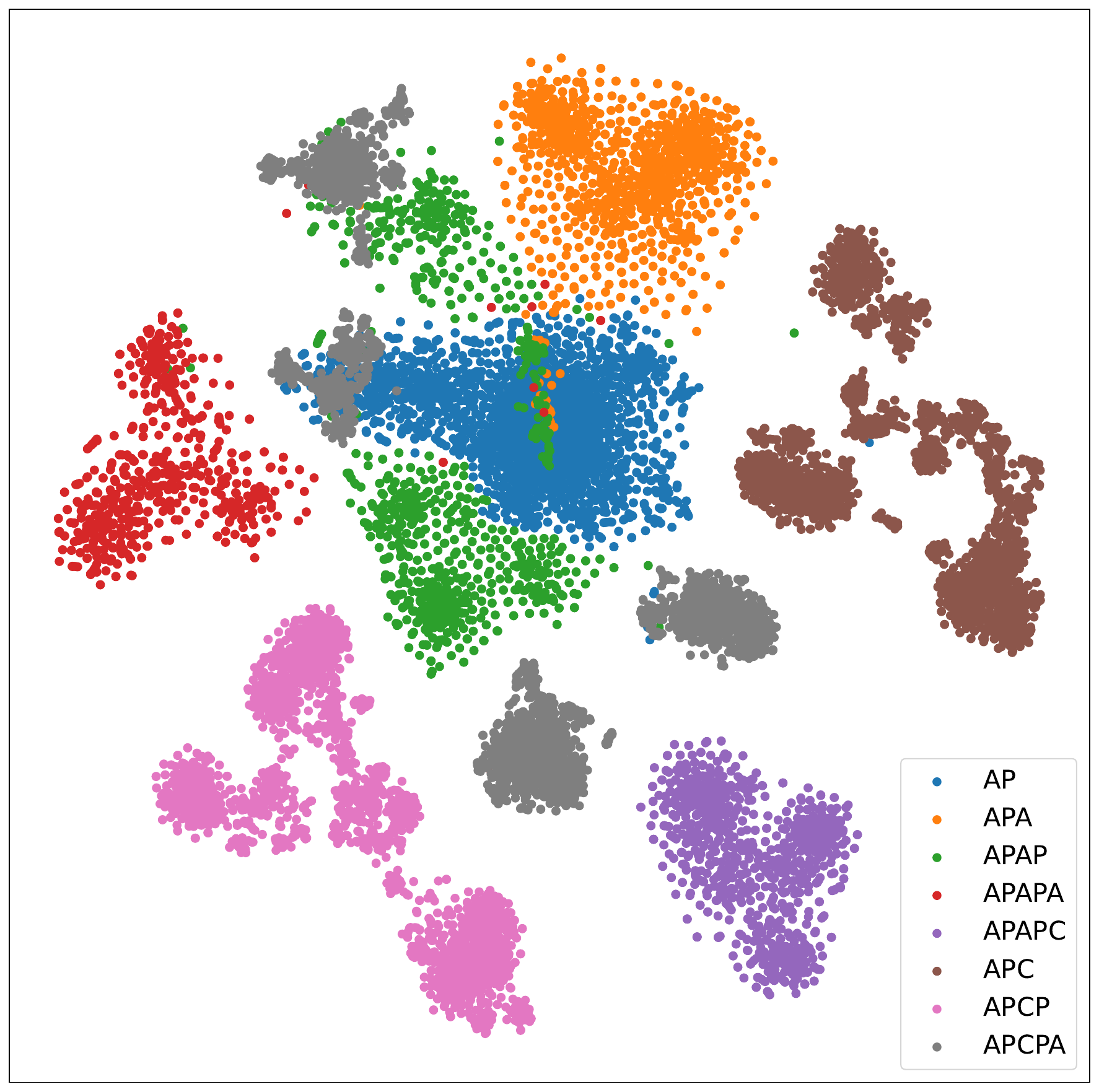}
            \subcaption{DBLP w/ \algname{}}%$\mathcal{L}_{hyp}$
        \end{minipage}\centering
        \begin{minipage}[b]{0.24\linewidth}
            \centering
            \includegraphics[width=\linewidth]{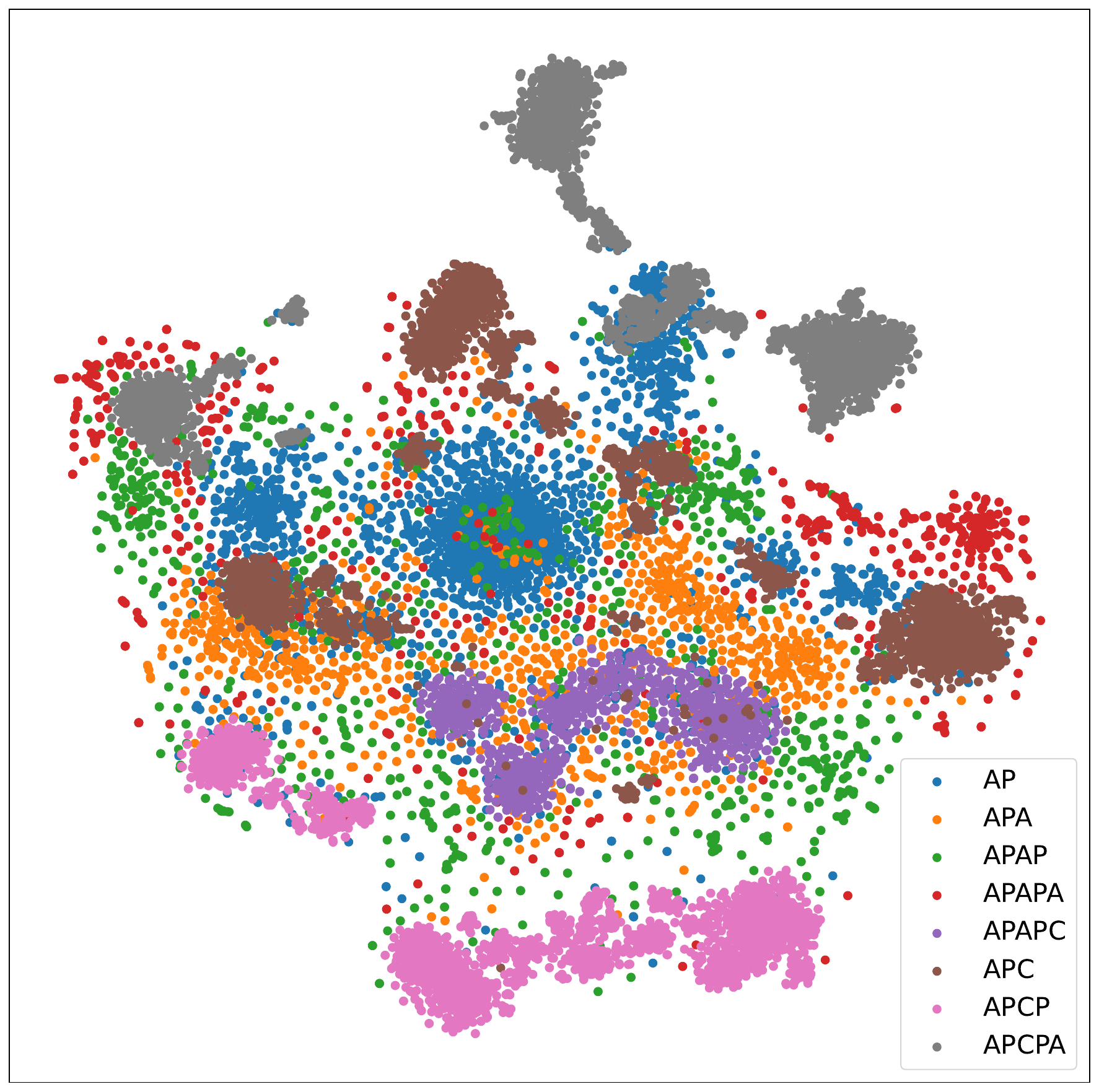}
            \subcaption{DBLP w/o \algname{}}%$\mathcal{L}_{hyp}$
        \end{minipage}
    \caption{Visualization of metapath embeddings on the ACM and DBLP datasets.}
\label{fig:visual}
\end{figure*}

\subsection{Visualization (RQ5)}
Metapath embeddings were visualized for an intuitive validation of the effectiveness of the proposed contrastive learning method in enhancing the semantic separation between embeddings derived from different metapaths. 
The training ratio was set to 40\% and metapath embeddings are projected onto a two-dimensional space using the t-SNE algorithm. 

Figures~\ref{fig:visual}(a) and (c) show metapath embeddings learned with the proposed contrastive learning method on the ACM and DBLP datasets, respectively, Figures~\ref{fig:visual}(b) and (d) show metapath embeddings learned without contrastive learning. For both datasets, we can observe a clear improvement in the separability of embeddings when the proposed contrastive learning method is applied.

Without the proposed contrastive learning method, embeddings from different metapaths (represented in different colors) tend to be mixed together, leading to poor discriminability. This makes it difficult to distinguish heterogeneous information derived from different metapaths. In contrast, with the proposed contrastive learning method, embeddings corresponding to each metapath form more compact and well-separated clusters. This indicates that \algname{} can capture the semantic and structural differences among metapaths. These results demonstrate that the metapath-based hyperbolic contrastive learning method enables \algname{} to learn more meaningful and distinct representations for each metapath and obtain enhanced heterogeneous graph representations.

\subsection{Complexity Analysis}
\begin{figure}[t!]
\captionsetup[subfigure]{justification=centering}
\centering
    \begin{minipage}[b]{0.49\linewidth}
        \centering
        \includegraphics[width=\linewidth]{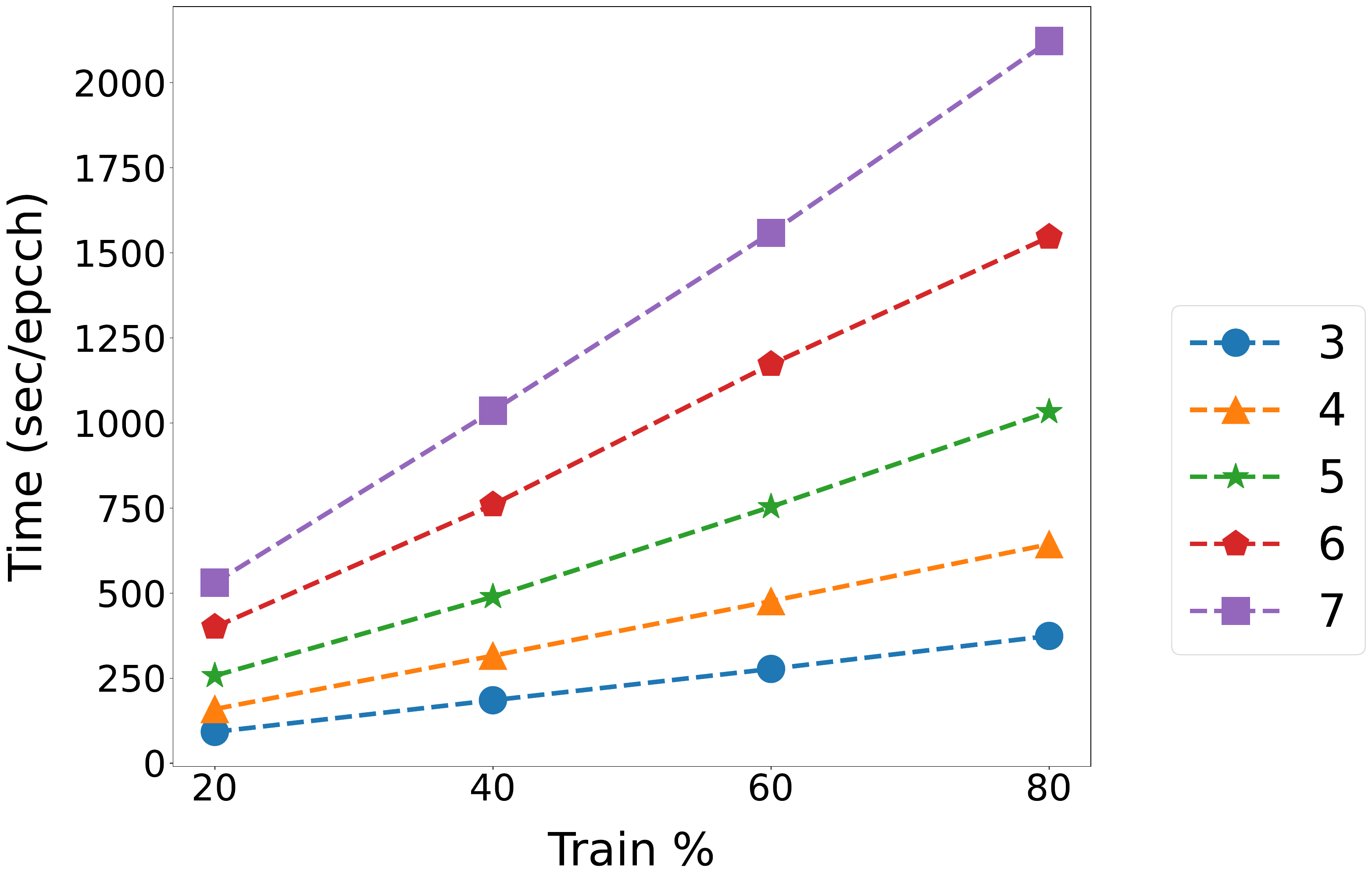}
        \subcaption{Train time according to the size of training sets.}
    \end{minipage}
    \begin{minipage}[b]{0.49\columnwidth}
        \centering
        \includegraphics[width=\linewidth]{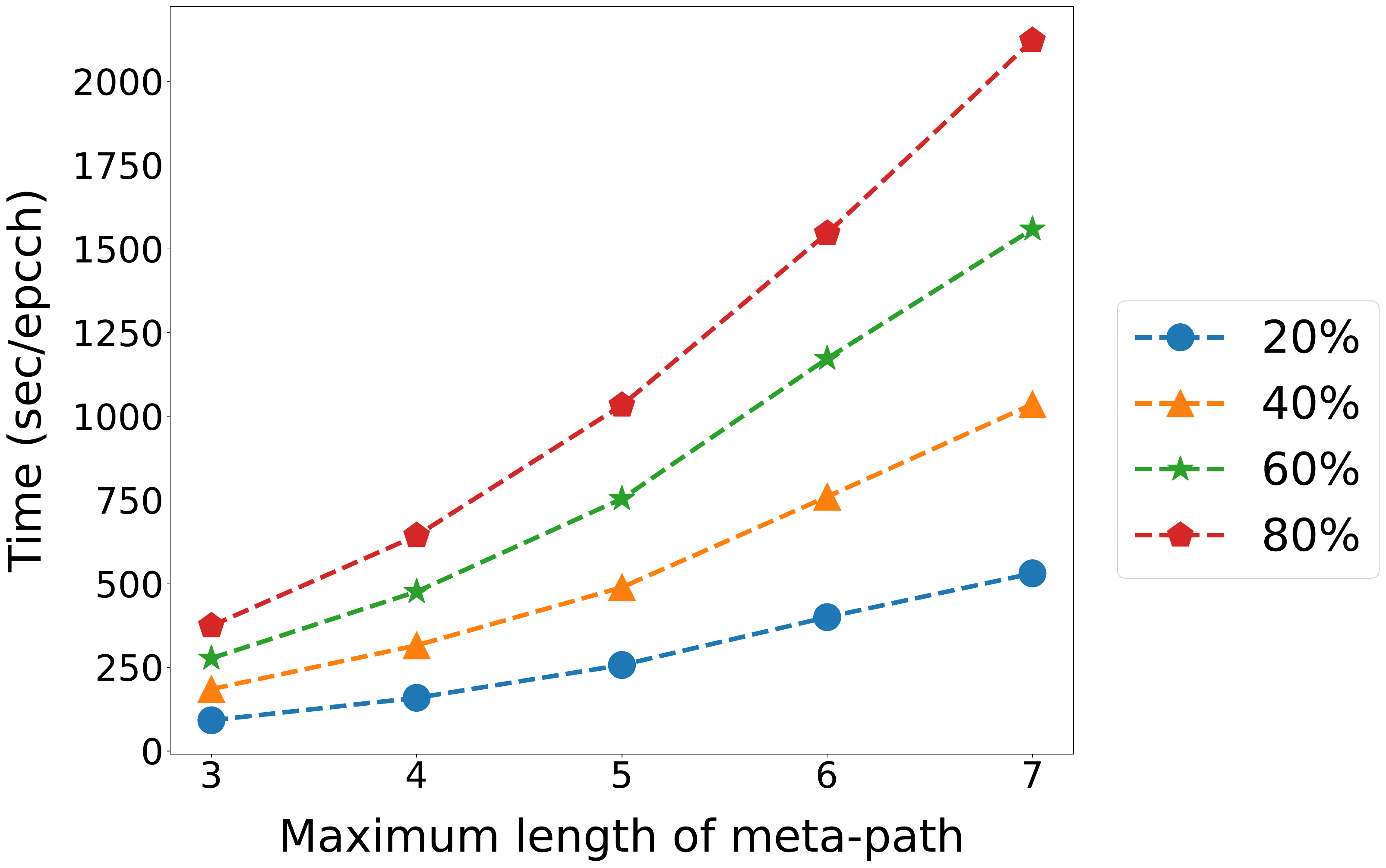}
        \subcaption{Train time according to hyperparameter $l$.}
    \end{minipage}
\caption{Training time of \algname{} on the ACM dataset.}
\label{figure:efficiency}
\end{figure}

Figure~\ref{figure:efficiency} illustrates the computational complexity for training of \algname{}. Specifically, in Figures~\ref{figure:efficiency}(a) and (b), each color represents the hyperparameter $l$ and the proportion of the training dataset, respectively. As theoretically shown in Section~\ref{sec:complexity}, the two main factors affecting the time complexity of \algname{} are the size of the dataset and the hyperparameter $l$, determining the maximum length of metapaths.  The training time of \algname{} increases approximately linearly with respect to both of these factors.

Figure~\ref{figure:efficiency}(a) shows the training time per epoch as the proportion of training data increases, for various values of $l$. Similarly, Figure~\ref{figure:efficiency}(b) shows the training time per epoch with increasing values of $l$, for different proportions of training data. Since the amount of training data increases, the number of nodes and links increases. Additonally, as hyperparameter $l$ increases, the number of sampled metapath and metapath instances also increase. As illustrated in Figure~\ref{figure:efficiency}, we empirically confirm that the training time per epoch increases linearly according to size of the training data and hyperparameter $l$, which aligns with our theoretical analysis in Section~\ref{sec:complexity}.

\section{Conclusion}
\label{sec:Conclusion}
In this paper, we proposed the \algname{} method that uses multiple hyperbolic spaces to effectively capture the distinct complex structures inherent in heterogeneous graphs. Furthermore, by adopting a metapath-based hyperbolic contrastive learning method, \algname{} learns the relationships between different metapaths and enhances the discriminability of heterogeneous semantic information derived from them.

We conducted comprehensive experiments to evaluate the effectiveness of \algname{} with widely used real-world heterogeneous graph datasets and synthetic heterogeneous graphs. The experiment results demonstrate that \algname{} consistently outperforms the other state-of-the-art baselines. Especially, compared to models such as MSGAT, which also utilize multiple hyperbolic spaces, \algname{} achieved superior performance by more effectively distinguishing differences among metapaths. Moreover, unlike homogeneous hyperbolic models such as HGCL, which are limited in modeling heterogeneous information, \algname{} can capture both the complex structures and heterogeneity of real-world graphs.

Our findings highlight the importance of contrastive learning in the hyperbolic space for learning heterogeneous graphs and suggest that integrating semantic guidance via metapaths can improve heterogeneous graph representation learning. For future work, we plan to enhance the computational efficiency of \algname{}. Although \algname{} effectively overcomes the need for domain-specific knowledge by sampling all metapaths within a maximum metapath length $l$ and learning importance for each, the number of sampled metapaths increases significantly as $l$ grows, leading to reduced computational efficiency. To address this issue, an automatic metapath selection procedure that focuses only on the most informative metapaths can be adopted.

%\algname{} samples all metapaths of length within $l$ and learns attention weights for each metapath, overcomes the limitation of requiring prior domain knowledge to define metapath. However, as $l$ increases, the number of sampled metapaths grows, leading to decreased computational efficiency. To address this issue, a promising future direction is to adopt an automatic metapath selection procedure that focuses only on the most informative metapaths, thereby improving computational efficiency.
\iffalse
\section*{Acknowledgement}
This work was supported by the National Research Foundation of Korea (NRF) grant funded by the Korea government (MSIT) (No.RS-2023-00214065) and by the Institute of Information \& Communications Technology Planning \& Evaluation (IITP) grant funded by the Korea government (MSIT) (No.RS-2022-00155857, Artificial Intelligence Convergence Innovation Human Resources Development (Chungnam National University)) and by the BK21 FOUR Program by Chungnam National University Research Grant, 2025.
\fi
\bibliographystyle{IEEEtran}
\bibliography{citations}

\end{document}